\documentclass[sigconf]{acmart}
\AtBeginDocument{%
  }

\setcopyright{cc}
\setcctype{by-nc}
\copyrightyear{2025}
\acmYear{2025}
\acmDOI{10.1145/XXXXXX.XXXXXX}
\acmConference[KDD '25] {Proceedings of the 31st ACM SIGKDD Conference on Knowledge Discovery and Data Mining V.2}{August 3--7, 2025}{Toronto, ON, Canada.}
\acmBooktitle{Proceedings of the 31st ACM SIGKDD Conference on Knowledge Discovery and Data Mining V.2 (KDD '25), August 3--7, 2025, Toronto, ON, Canada}
\acmISBN{979-8-4007-1454-2/25/08}
\acmDOI{10.1145/3711896.3736876}

\newcommand{\ours}{CellCLAT}
\usepackage{booktabs} 
\usepackage{multirow}
\usepackage{algorithm}
\usepackage{algorithmic}
\newtheorem{definition}{Definition}

\newtheorem{theorem}{Theorem}
\newtheorem{lemma}{Lemma}

\usepackage{MnSymbol}
\newcommand{\attention}[1]{\textbf{\textcolor{red}{***}}}

\begin{document}

\title{CellCLAT: Preserving Topology and Trimming Redundancy in Self-Supervised Cellular Contrastive Learning}

\author{Bin Qin}
\authornote{All authors contributed equally to this research.}
\email{qinbin21@mails.ucas.ac.cn}
\orcid{0009-0001-9023-0293}
\affiliation{%
  \institution{Institute of Software, Chinese Academy of Sciences}
  \city{Beijing}
  \country{China}
}
\affiliation{%
  \institution{University of Chinese Academy of Sciences}
  \city{Beijing}
  \country{China}
}

\author{Qirui Ji}
\authornotemark[1]
\email{jiqirui2022@iscas.ac.cn}
\orcid{0009-0006-5982-9739}
\affiliation{%
  \institution{Institute of Software, Chinese Academy of Sciences}
  \city{Beijing}
  \country{China}
}
\affiliation{%
  \institution{University of Chinese Academy of Sciences}
  \city{Beijing}
  \country{China}
}

\author{Jiangmeng Li}
\authornote{Corresponding Author.}
\authornotemark[1]
\email{jiangmeng2019@iscas.ac.cn}
\orcid{0000-0002-3376-1522}
\affiliation{%
  \institution{Institute of Software, Chinese Academy of Sciences}
  \city{Beijing}
  \country{China}
}
\affiliation{%
  \institution{University of Chinese Academy of Sciences}
  \city{Beijing}
  \country{China}
}

\author{Yupeng Wang}
\email{yupeng@iscas.ac.cn}
\orcid{0009-0008-3462-0335}
\affiliation{%
  \institution{Institute of Software, Chinese Academy of Sciences}
  \city{Beijing}
  \country{China}
}

\author{Xuesong Wu}
\authornotemark[2]
\email{xuesong@iscas.ac.cn}
\orcid{0000-0002-9534-6906}
\affiliation{%
  \institution{Institute of Software, Chinese Academy of Sciences}
  \city{Beijing}
  \country{China}
}

\author{Jianwen Cao}
\authornotemark[2]
\email{jianwen@iscas.ac.cn}
\orcid{0000-0003-4329-3790}
\affiliation{%
  \institution{Institute of Software, Chinese Academy of Sciences}
  \city{Beijing}
  \country{China}
}
\affiliation{%
  \institution{University of Chinese Academy of Sciences}
  \city{Beijing}
  \country{China}
}

\author{Fanjiang Xu}
\email{fanjiang@iscas.ac.cn}
\orcid{0000-0002-4981-6217}
\affiliation{%
  \institution{Institute of Software, Chinese Academy of Sciences}
  \city{Beijing}
  \country{China}
}
\affiliation{%
  \institution{University of Chinese Academy of Sciences}
  \city{Beijing}
  \country{China}
}

\renewcommand{\shortauthors}{Bin Qin et al.}

\begin{abstract}
   Self-supervised topological deep learning (TDL) represents a nascent but underexplored area with significant potential for modeling higher-order interactions in simplicial complexes and cellular complexes to derive representations of unlabeled graphs. Compared to simplicial complexes, cellular complexes exhibit greater expressive power. However, the advancement in self-supervised learning for cellular TDL is largely hindered by two core challenges: \textit{extrinsic structural constraints} inherent to cellular complexes, and \textit{intrinsic semantic redundancy} in cellular representations. The first challenge highlights that traditional graph augmentation techniques may compromise the integrity of higher-order cellular interactions, while the second underscores that topological redundancy in cellular complexes potentially diminish task-relevant information. To address these issues, we introduce \textit{\textbf{Cell}ular Complex \textbf{C}ontrastive \textbf{L}earning with \textbf{A}daptive \textbf{T}rimming} (CellCLAT), a twofold framework designed to adhere to the combinatorial constraints of cellular complexes while mitigating informational redundancy. Specifically, we propose a parameter perturbation-based augmentation method that injects controlled noise into cellular interactions without altering the underlying cellular structures, thereby preserving cellular topology during contrastive learning. Additionally, a cellular trimming scheduler is employed to mask gradient contributions from task-irrelevant cells through a bi-level meta-learning approach, effectively removing redundant topological elements while maintaining critical higher-order semantics. We provide theoretical justification and empirical validation to demonstrate that CellCLAT achieves substantial improvements over existing self-supervised graph learning methods, marking a significant attempt in this domain. 
\end{abstract}




\maketitle

\newcommand\kddavailabilityurl{https://doi.org/xxxx}

\ifdefempty{\kddavailabilityurl}{}{
\begingroup\small\noindent\raggedright\textbf{KDD Availability Link:}\\
The source code of this paper has been made publicly available at \url{https://doi.org/10.5281/zenodo.15514849}.
\endgroup
}

\section{Introduction}
Graph Neural Networks (GNNs) \cite{scarselli2008graph} model \textit{pairwise interactions} in non-Euclidean graph structures, such as user relationship modeling in social networks \cite{fan2019graph}, protein-protein interactions \cite{rao2014protein} in biology, and chemical or molecular property prediction \cite{wu2018moleculenet}. GNNs achieve this by aggregating the $k$-hop neighborhood information, which follows the \textit{graph topology}, to learn node representations. However, growing demands for \textit{higher-order interaction} modeling have motivated emerging researches \cite{papamarkou2024position} to design more expressive GNNs \cite{morrisposition} capable of identifying fine-grained topological patterns, encoding substructures, and capturing long-range dependencies. For instance, protein complex identification \cite{zahiri2020protein} requires group-wise interaction analysis, and social network group behavior \cite{arya2018exploiting} modeling necessitates representations beyond conventional graph structures. These challenges reveal the insufficiency of graphs in expressing complex \textit{data relations} with multifaceted interactions. Topological deep learning (TDL) has consequently emerged, shifting the focus from graphs to \textit{combinatorial topological spaces} that learn representations of simplicial complexes or cellular complexes \cite{edelsbrunner2010computational}. The message-passing mechanism-based neural networks within these spaces demonstrate expressive capabilities that are at least equivalent to those of the 3-Weisfeiler-Lehman (3-WL) test \cite{bodnar2021weisfeiler2, eitan2024topological}. In contrast to traditional node-wise message-passing, TDL enables the aggregation of messages through multi-level relationships among simplices or cells, effectively capturing higher-order interactions.

While hierarchical high-order interactions are highly advantageous, high-dimensional topological objects such as simplicial or cellular complexes present significant challenges in terms of data annotation and task definition compared to standard graph structures. These challenges arise from their inherently complex high-dimensional topological structures (e.g., defining cliques or cycles within a graph), their reliance on domain-specific knowledge, and the presence of topological features that do not directly correspond to conventional labels (e.g., persistent homology and topological invariants). As a result, there is an urgent need for developing self-supervised TDL. Pioneering studies  \cite{madhu2024toposrl} have utilized contrastive learning paradigms to learn representations from simplicial complexes. However, as a more expressive topological structure, cellular complex-based self-supervised TDL approaches remains largely underexplored. This gap primarily stems from the fact that existing graph-based self-supervised learning (SSL) techniques are not directly applicable, due to two fundamental challenges: \textbf{extrinsic structural constraints} inherent in cellular complexes, and \textbf{intrinsic semantic redundancy} in cellular representations.

\begin{figure*}
    \centering
    \includegraphics[width=1.0\linewidth,alt={Motivation.}]{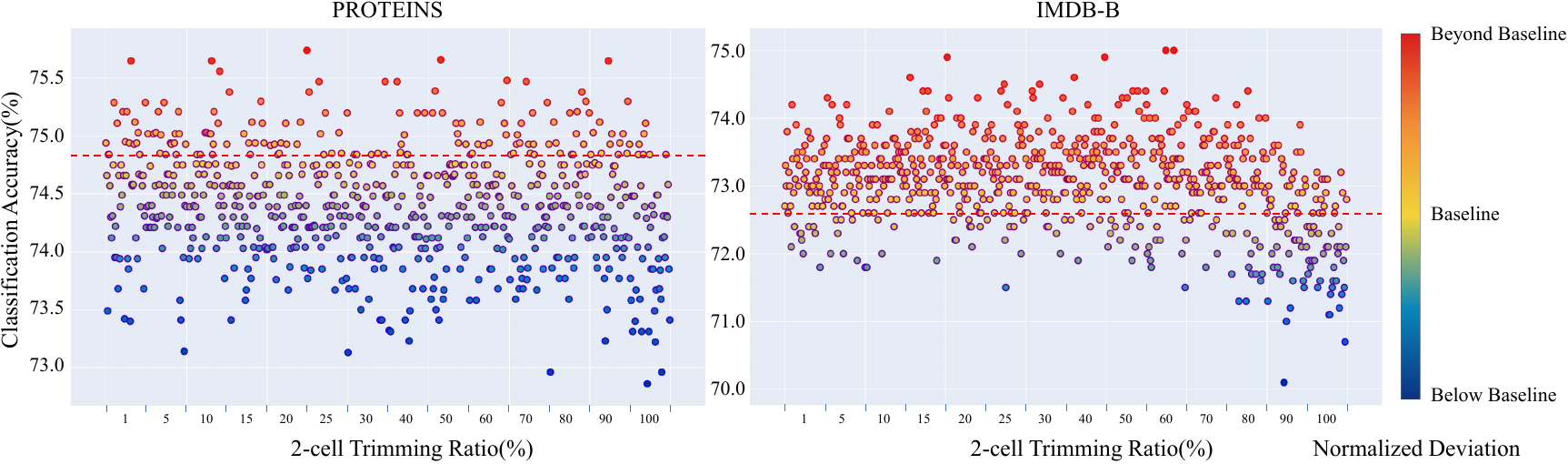}
    \caption{Experimental scatter diagrams obtained by randomly trimming 2-cell cellular complex contrastive learning representations on the PROTEINS and IMDB-B datasets. The baseline and the red dashed line indicate the classification accuracy achieved using the complete 2-cell representations. The x-axis values represent a fixed proportion of 2-cell removal, with each proportion interval containing 50 points. Each individual point corresponds to an independent classification result achieved by randomly trimming the original 2-cell representations with a specific trimming ratio.}
    \label{fig:motivation}
    \vspace{-0.32cm}
\end{figure*}

The predominant augmentation technique in graph SSL involves directly applying node-dropping or edge-perturbation to graphs \cite{graphcl}. However, such "loose" perturbations are insufficient for generating augmented views of cellular complexes while preserving their global topological properties. Accordingly, structural constraints \cite{whitehead1949combinatorial} refer to the \textit{continuity of attaching maps} and the \textit{closure-finiteness} in the \textit{gluing process} for constructing cellular complexes \cite{hansen2019toward}. For instance, if a 2-cell (modeled by a polygon) has a boundary that includes a deleted edge $e$, the attaching maps of that 2-cell become invalid. Similarly, if a node $v$ is removed, all 1-cells (edges) incident to $v$ must also be removed; otherwise, closure-finiteness is violated. Despite adhering to such structural constraints, achieving augmented views that preserve semantically rich higher-order topological structures remains a significant challenge.

Another counterintuitive yet focal phenomenon is that not all higher-order cellular interactions carry task-relevant information. Some interactions may introduce semantic redundancy, which we term \textit{Cellular Topological Redundancy}. As depicted in the Figure \ref{fig:motivation}, we conduct exploratory experiments to validate this phenomenon by introducing a \textit{naive} cellular complex contrastive learning paradigm\footnote{We adopt a parameter perturbation-based augmentation method, which preserves the structural integrity of the input cellular complex while generating augmented views for contrastive learning, to obtain representations of cells at all orders within the cellular complex.}. Specifically, we observe that certain trimmed representations can indeed achieve better performance than the original baseline representation, providing evidence that trimming redundant cells can indeed improve performance.

To cope with the challenges, we propose \textit{Cellular Complex Contrastive Learning with Adaptive Trimming} (CellCLAT), a novel yet effective two-pronged framework designed to simultaneously preserve cellular topology and eliminate semantic redundancy within cellular complexes. The core innovation of our approach lies in two key components: 1) We develop a parameter perturbation-based augmentation method that injects \textit{controlled} noise into the cellular interaction process, rather than directly altering the cellular structures, to maximize the preservation of cellular topology during contrastive learning. This component circumvents the destruction of extrinsic structural constraints (e.g., attaching map continuity and closure-finiteness properties) inherent to cellular complexes during generating contrastive views. 2) Based on the obtained comprehensive cellular representations obtained, we propose a \textit{cellular trimming scheduler} that adaptively trims redundant 2-cells in a self-paced manner. Concretely, the scheduler employs a bi-level meta-learning optimization objective that integrates the trimming process with contrastive learning tasks, enabling the framework to dynamically block gradient contributions from topological elements containing confounding or task-irrelevant information. As a result, superfluous higher-order structures are discarded, enhancing the focus on task-relevant topological patterns.

Our \textbf{contributions} are summarized as follows: 1) To the best of our belief, we present the \textit{first} self-supervised method for Cellular TDL, effectively adhering the extrinsic structural constraints inherent to cellular complexes. 2) We conduct an in-depth investigation into the semantic redundancy of higher-order cellular topology and propose an adaptive cellular trimming scheduler to explore the higher-order interactions with task-relevant semantics. 3) We leverage the cellular Weisfeiler-Lehman test to demonstrate that CellCLAT‘s expressiveness surpasses GNN-based SSL methods in distinguishing non-isomorphic graphs, and we further provide solid theoretical justification from a causal perspective to determine that the cellular topological redundancy is treated as a confounder. 4) We conduct extensive experiments across various benchmarks to empirically validate the superior performance of CellCLAT.

\section{Related Work}
\subsection{Topological Deep Learning}
TDL extends the paradigm of Message Passing Neural Networks (MPNNs) \cite{gilmer2017neural} by generalizing message passing beyond graphs to relational data embedded in combinatorial topological spaces. Such relational data \cite{papamarkou2024position} often exhibit higher-order structures that are inadequately captured by traditional graph-based methods, leading to the development of two primary architectures: \textbf{Message Passing Simplicial Networks (MPSNs)} \cite{bodnar2021weisfeiler1,ebli2020simplicial,roddenberry2021principled,wu2023simplicial} and \textbf{Message Passing Cellular Networks (MPCNs)} \cite{hajij2020cell, bodnar2021weisfeiler2, giusti2023cin++}. MPSNs extend the Weisfeiler-Lehman (WL) isomorphism test \cite{weisfeiler1968reduction} to simplicial complexes (SCs) through a hierarchical coloring procedure \cite{bodnar2021weisfeiler1}, proving strictly more expressive than graph-based MPNNs. MPCNs further advance this by operating on cellular complexes (CWs), topological objects that flexibly generalize both SCs and graphs, thereby overcoming the strict combinatorial constraints (e.g., only cliques in graph can be referred to as simplices) inherent to SCs \cite{bodnar2021weisfeiler2}. Moreover, the recent development of combinatorial complexes (CCs) provides a more general framework that unifies SCs, CWs, and hypergraphs, enabling more flexible message-passing architectures \cite{hajij2022higher, papillon2024topotune, battiloro2024n}. 

The landscape of TDL architectures has diversified through convolutional and attentional mechanisms tailored to topological domains. \textbf{Convolutional designs} for simplicial \cite{yang2022efficient, yang2023convolutional, yang2022simplicial, yang2022simplicial1} and cellular complexes have proliferated, demonstrating how localized filters or kernels can be defined over higher-order simplices and cells to enhance expressivity. Parallel efforts have explored \textbf{attentional mechanisms}, adapting self-attention layers to simplicial complexes \cite{giusti2022simplicial, battiloro2024generalized, goh2022simplicial, lee2022sgat} and more general cellular \cite{giusti2023cell} or combinatorial complexes \cite{hajij2022higher}. These attention-based networks often re-weight interactions between neighboring simplices or cells in orientation-equivariant ways \cite{goh2022simplicial}, while also considering both upper and lower neighborhoods for feature aggregation \cite{giusti2022simplicial}. Beyond straightforward higher-order message passing, alternative approaches incorporate pre-computed topological information—e.g., persistent homology \cite{ballester2023expressivity, verma2024topological} or other invariants \cite{buffelli2024cliqueph, dan2024tfgda}—into graph-level or cell-level features, thereby enriching the learned representations \cite{horn2021topological, chen2021topological}. Comprehensive surveys \cite{papillon2304architectures, papamarkou2024position, eitan2024topological} underscore the potential of TDL to address complex data modalities where higher-order and geometric structures play an important role. However, the literature on SSL in TDL remains scarce, with only limited pioneering works \cite{madhu2024toposrl} proposed for simplicial complexes to date.

\subsection{Graph Contrastive Learning}
Numerous methods have been explored to advance graph-level contrastive learning \cite{graphcl, joao, adgcl, rgcl, simgrace, html,drgcl}, with a primary focus on designing effective data augmentation techniques and contrastive objectives to enhance graph representation learning.
GraphCL \cite{graphcl} introduces four general types of augmentations to enforce consistency between different transformed graph views. ADGCL \cite{adgcl} employs adversarial graph augmentation strategies to reduce the risk of capturing redundant information. JOAO \cite{joao} addresses the challenge of selecting appropriate augmentations by proposing a min-max two-level optimization framework based on GraphCL. RGCL \cite{rgcl} incorporates invariant rationale discovery to generate robust graph augmentations, ensuring that the structural rationale is preserved in augmented views. SimGRACE \cite{simgrace} eliminates the need for extensive augmentation search by perturbing the encoder with random Gaussian noise, offering a more efficient approach to generating augmented views. 
HTML \cite{html} employs knowledge distillation by incorporating the learning of graph-level and subgraph-level topological isomorphism tasks into the objective function, thereby enhancing the performance of downstream tasks.
Our method introduces cellular complexes into graph contrastive learning, while investigating the cellular topological redundancy issue.

\section{Methodology}
\label{framework}
\begin{figure*}
    \centering
    \includegraphics[width=0.95\linewidth,alt={Framework.}]{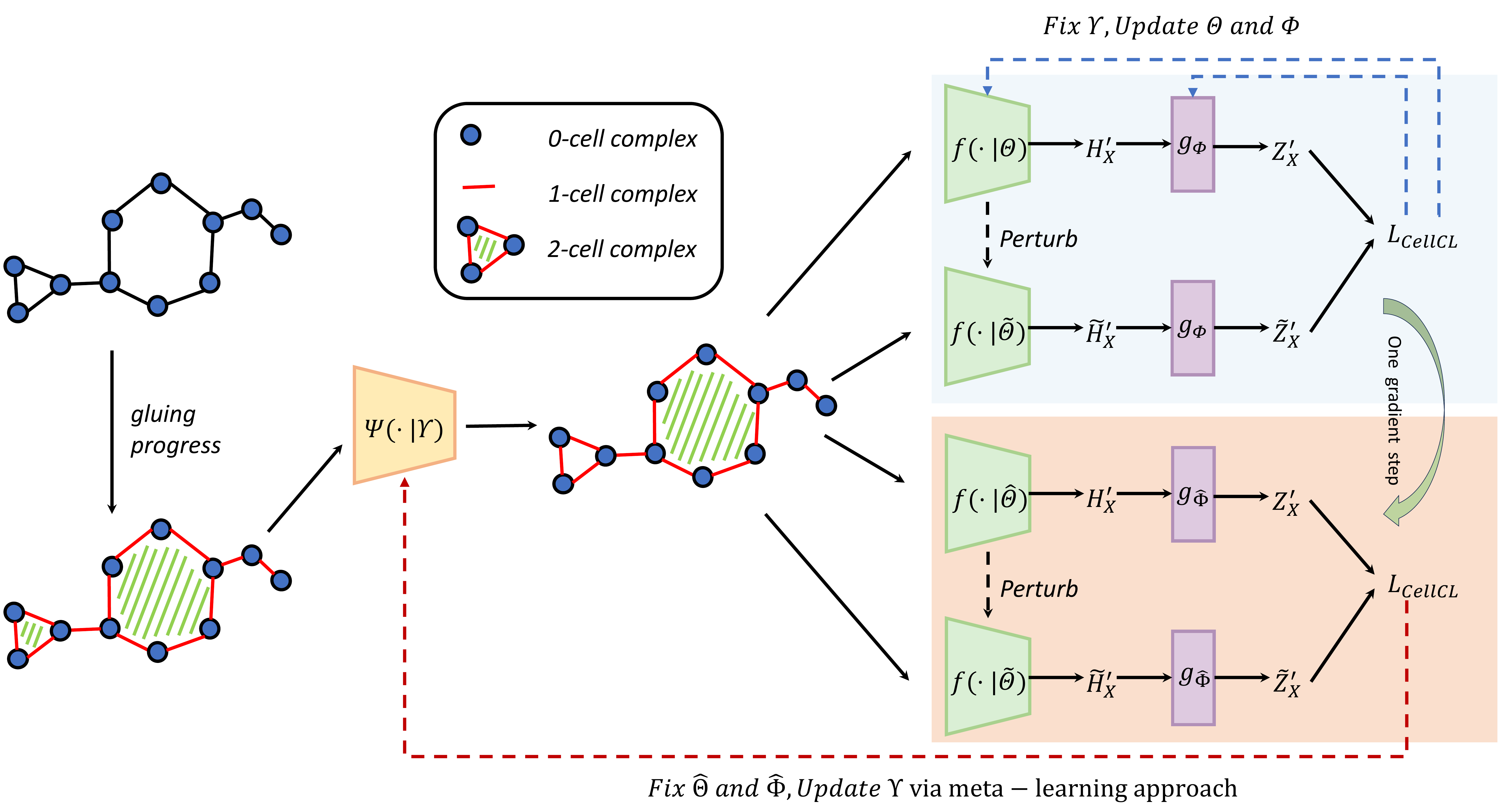}
    \caption{The framework of $\text{CellCLAT}$. The blue dashed lines indicate the standard contrastive learning phase, where the encoder $f(\cdot;\Theta)$ and the projection head $g(\cdot;\Phi)$ are updated while keeping the Cellular Trimming Scheduler $\Psi$ fixed. The red dashed lines represent the update process of $\Psi$ through the bi-level optimization process.}
    \label{fig:framework}
\end{figure*}

In this section, we introduce a feasible self-supervised solution that is adaptable to the current topological deep learning, as illustrated in Figure \ref{fig:framework}. Specifically, in Section \ref{CCNN}, we summarize the current learning paradigm of cellular complex networks, which extends the strictly restricted combinatorial structure of simplicial complexes. It not only inherits the structural properties of simplicial complexes but also introduces additional flexibility, enabling the handling of more complex and flexible topological structures. \footnote{In the two-dimensional case, simplicial complexes are limited to triangular simplices, whereas two-dimensional cellular complexes can represent arbitrary induced cycles.} We develop a self-supervised TDL framework based on cellular complex networks, which exhibits a stronger topological expressiveness compared to GNN-based models. In Section \ref{contrastive}, we implement a self-supervised framework that maximizes the retention of cellular topology of the original cellular complex. By introducing perturbations (such as Gaussian noise) at the encoder level, rather than directly augmenting the input data, we can avoid damaging the local cellular structures of the cellular complex. This approach allows us to generate the necessary different views for contrastive learning while preserving the integrity of higher-order cellular topology. Finally, in the Section \ref{redundant}, we explore the semantically relevant components within cellular topology. A cellular trimming scheduler trims the complete cellular representations obtained in Section \ref{contrastive} and employs a bi-level meta-learning optimization process to suppress the gradient contributions of task-irrelevant cells, thereby achieving a dynamic balance between the need for higher-order interactions and task-relevant semantics.

 \begin{figure}
   \centering
     \includegraphics[width=0.9\linewidth,alt={Gluing.}]{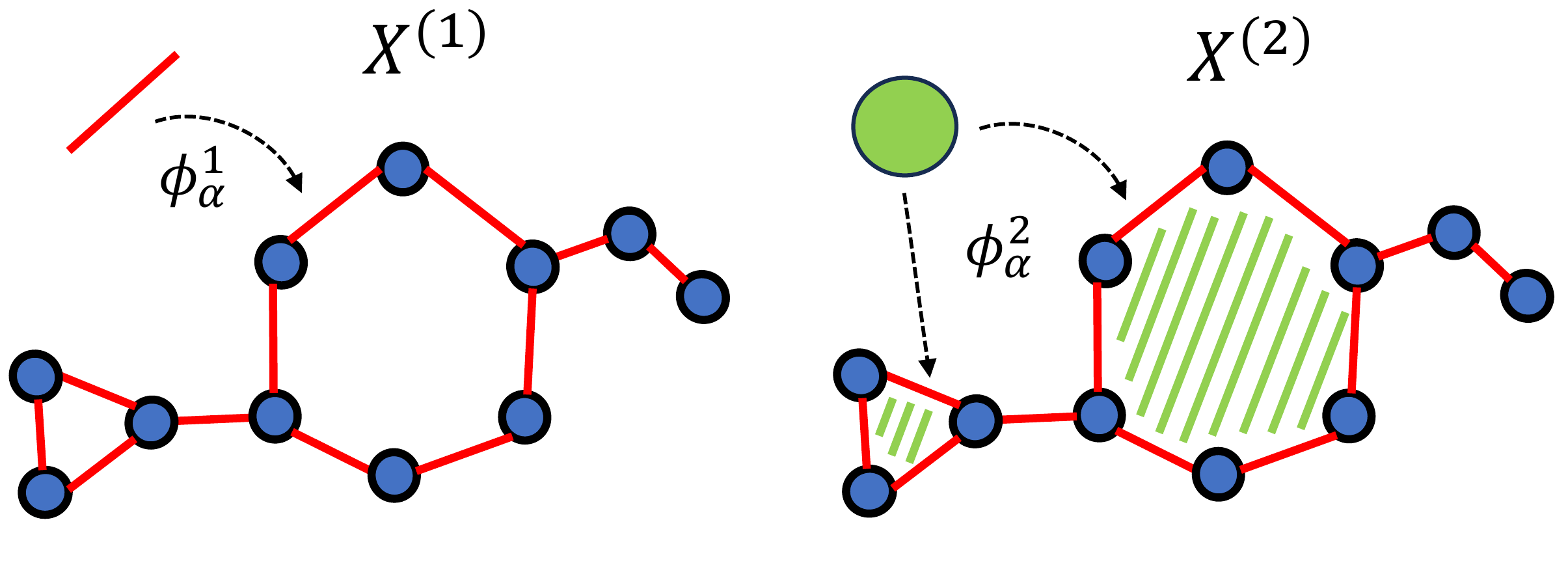}
     \caption{The gluing process of constructing a cellular complex from a graph is achieved through a sequence of continuous attaching maps.}
     \label{fig:gluing}
 \end{figure}

\subsection{Learning Paradigm of Cellular Complex Neural Networks}
\label{CCNN}
Topological deep learning focuses on topological spaces formed by structured data in non-Euclidean spaces, which can leverage the rich geometric and topological properties within the data. Here, we primarily focus on cellular complexes. A cellular complex is a topological space built from a collection of \textbf{cells} of various dimensions, composed of discrete sets of points. An $n$-dimensional cell, denoted as $n$-cell, is a space homeomorphic to an open $n$-dimensional ball $B^n(x,r)=\left\{ x\in \mathbb{R}^n:\left\| x \right\|< r \right\}$. Formally, the construction of a cellular complex is achieved through a recursive gluing process:

\begin{enumerate}
    \item 0-Skeleton: Start with a discrete set of points, $X^{(0)}$, which serves as the \textbf{0-skeleton} of the complex.
    \item Attaching Higher-Dimensional Cells: To construct the $n$-skeleton $X^{(n)}$, attach $n$-cells $\{\sigma^{(n)}_{\alpha}\}$ to $X^{(n-1)}$ via continuous maps called \textbf{attaching maps}: $\phi_\alpha^n:\partial \sigma_\alpha^{(n)}\to X^{(n-1)}.$ The new space $X^{(n)}$ is defined as the union: $X^{(n)}=X^{(n-1)}\cup\bigcup_\alpha \sigma^{(n)}_{\alpha},$ where each $\sigma^{(n)}_{\alpha}$ is glued to $X^{(n-1)}$ along its boundary using $\phi_\alpha^n$.
\end{enumerate}

Taking the graph $G=(V,E)$ as an example, we provide an intuitive explanation of how the cellular complex $X$ extends the topological structure of the graph, as shown in Figure \ref{fig:gluing}. Starting from the vertex set $V$, the 0-cells in $X$ correspond to these vertices, i.e., $V = X^{(0)}$. Next, by gluing the endpoints of line segments to these vertices, we obtain the 1-cells $\{\sigma^{(1)}_{\alpha}\}$ in $X$, which correspond to the edges $E$. Therefore, the graph $G$ can be seen as the one-dimensional skeleton $X^{(1)}$ of the cellular complex $X$. Finally, take a two-dimensional closed disk and attach its boundary (a circle) to any (induced) cycle in graph $G$. Hence, \textbf{polygons} in the graph can be represented as 2-cells, with their boundaries being the edges that form the cycle. The 2-dimensional cellular complex $X = X^{(2)}$ characterizes higher-order relational structures while restricting modifications to the original graph structure.

Unlike graphs, which only have a single adjacency relationship, higher-order interactions in cellular complexes are manifested in the \textbf{neighborhood structure} defined between cells of different dimensions \cite{papillon2304architectures, hajij2206topological}. Let $\sigma \prec \tau$ denote that the $(r-1)$-cell $\sigma$ is the \textbf{boundary} of the $r$-cell $\tau$, for example, nodes are the boundaries of edges, and edges are the boundaries of polygons. A $r$-cell $\tau$ has four types of neighborhood structures \cite{bodnar2021weisfeiler1, bodnar2021weisfeiler2}: Boundary Adjacent Neighborhood $\mathcal{B}(\tau)=\left\{ \sigma \mid \sigma \prec \tau \right\}$, Co-Boundary Adjacent Neighborhood $\mathcal{C}(\tau)=\left\{ \sigma \mid \tau \prec \sigma \right\}$, Lower Adjacent Neighborhood $\mathcal{N}_{\downarrow}(\tau)=\left\{ \sigma \mid \exists \ \delta \ \text{s.t.} \  \delta \prec \sigma \wedge \delta \prec \tau\right\}$, Upper Adjacent Neighborhood $\mathcal{N}_{\uparrow}(\tau)=\left\{ \sigma \mid \exists \ \delta \ \text{s.t.} \ \sigma \prec \delta\wedge \tau \prec \delta \right\}$.

A \textbf{Cellular Complex Neural Network (CCNN)} built upon the message passing paradigm updates features associated with cells in a cellular complex through hierarchical, structured interactions across cells of varying dimensions. Given a $r$-cell $\tau$ and $r'$-cell $\sigma$, let the embedding at the $l$-th layer be denoted as $h^{(l)}_r(\tau)$ and $h^{(l)}_{r'}(\sigma)$. First, the message received by $\tau$ is computed as $m(\sigma \rightarrow \tau) = \psi_{\mathcal{N}_i} ( h^{(l)}_r(\tau), h^{(l)}_{r'}(\sigma), \Theta_i^{(l)} )$, where $\mathcal{N}_i\in \mathcal{N}=\left\{ \mathcal{B}, \mathcal{C}, \mathcal{N}_{\downarrow}, \mathcal{N}_{\uparrow}\right\}$ represents messages from four types of neighbors, and $\psi_{\mathcal{N}_i}$ denotes differentiable functions, such as MLPs, with $\Theta_i^{(l)}$ being the learnable parameters. Then, the messages undergo two types of aggregation: 1) intra-neighborhood aggregation $M_{\mathcal{N}_i}(\tau) = \bigoplus_{\sigma \in \mathcal{N}_i(\tau)} m(\sigma \rightarrow \tau)$, where $\bigoplus$ is a permutation-invariant operation (e.g. summation or mean); 2) inter-neighborh-ood aggregation $M(\tau) = \bigotimes_{\mathcal{N}_i \in \mathcal{N}} M_{\mathcal{N}_i}(\tau)$, where $\bigotimes$ is a combining function (e.g., summation or concatenation) that combines information from neighborhoods. The update of the embedding at the $(l+1)$-th layer can be formalized as:
\begin{equation}
    h^{(l+1)}_r(\tau)=\phi \left ( h^{(l)}_r(\tau), \bigotimes_{\mathcal{N}_i \in \mathcal{N}} \bigoplus_{\sigma \in \mathcal{N}_i(\tau)} \psi_{\mathcal{N}_i} \left( h^{(l)}_r(\tau), h^{(l)}_{r'}(\sigma), \Theta_i^{(l)} \right) \right ), \label{embedding}
\end{equation}
where $\phi$ is a learnable update function.

\subsection{Contrastive Learning with Maximized Cellular Topology Preservation}
\label{contrastive}
The complex topological structure of cellular complexes renders commonly used techniques for graph augmentation ineffective, as the unrestricted removal of nodes or edges may disrupt the continuity of the attaching map \cite{hansen2019toward,whitehead1949combinatorial}, making it impossible to reasonably define high-dimensional cells. \footnote{For instance, in the case of a 2-dimensional quadrilateral cell, if one of its edges is removed, the quadrilateral is no longer well-defined. The boundary of the quadrilateral (a topological circle) cannot be mapped to a disconnected path in the 1-skeleton.} TopoSRL \cite{madhu2024toposrl} is the first attempt to implement data augmentation for simplicial complexes. However, the random removal of closed simplices similarly violates the subset-closed property of higher-order simplices. Furthermore, the high-order structures in cellular complexes (e.g., polygons represented by 2-cells) often carry important semantic information, such as ring structures in molecular graphs or community structures in social networks. Random augmentation strategies, such as the addition or removal of polygons, may alter the semantics of these higher-order structures. Adaptive augmentation that preserves semantic information often requires manually selecting augmentation strategies for each dataset or performing a tedious search for suitable augmentation combinations, sometimes relying on expensive domain knowledge. These limitations become even more problematic when applied to cellular complex data structures.

To address this, inspired by SimGRACE \cite{simgrace}, we propose generating augmented embeddings for cellular complexes by introducing noise into the network parameters of the message-passing process. This approach avoids disrupting local topological structures, preserves high-order semantic information, and maximally extracts the topological information from the input views. Specifically, we implement Equation \eqref{embedding} as follows:
\begin{equation}
    \resizebox{.99\linewidth}{!}{$
            \displaystyle
            h^{(l+1)}_r(\tau)=\text{MLP}^{(l)}_{U,r}( \underset{\mathcal{N}_i \in \mathcal{N}}{\big\|} \text{MLP}^{(l)}_{\mathcal{N}_i,r}( (1+\epsilon_{\mathcal{N}_i} )h^{(l)}_r(\tau)+\sum_{\sigma \in \mathcal{N}_i(\tau)} h^{(l)}_{r'}(\sigma)))
        $}.
\end{equation}

The final embedding of the cellular complex $X$ is defined as $\mathbf{H}_{X}=\sum_{r=0}^{2}\text{MLP}_{X,r}(\sum h^{(L)}_r(\tau))$. Let $\text{MLP}^{(l)}_{U,r}$, $\text{MLP}^{(l)}_{\mathcal{N}_i,r}$, and $\text{MLP}_{X,r}$ have parameters $\Theta_{k,r}^{(l)}, k=U, \mathcal{N}_i, X$, respectively. We denote the corresponding encoder by $f_{\Theta}(\cdot)$. Gaussian noise is added to all parameters to obtain perturbed network parameters $\tilde{\Theta}_{k,r}^{(l)}=\Theta_{k,r}^{(l)}+\eta \cdot \varepsilon_{k,r}^{(l)},\text{where} \ \varepsilon_{k,r}^{(l)}\sim \mathcal{N}\left ( 0,(\text{std}(\Theta_{k,r}^{(l)}))^2 \right )$. Here, $\eta$ controls the magnitude of noise. Consequently, $\mathbf{H}_{X}$ and the augmented embedding $\tilde{\mathbf{H}}_X$ obtained from the perturbed network parameters constitute a positive pair. Suppose a batch contains $N$ cellular complexes $X_1, X_2, ..., X_N$, yielding a total of $2N$ embeddings in the latent space. Each cellular complex $X_i$ and its augmented embedding form a positive sample pair $(\mathbf{H}_{X_i},\tilde{\mathbf{H}}_{X_i})$, while augmented embeddings of other cellular complexes $X_j$ form negative sample pairs $(\mathbf{H}_{X_i},\tilde{\mathbf{H}}_{X_j})$.

To better align positive pairs and uniformly distribute negative pairs, as explored in \cite{wang2020understanding}, we project the embedding $\mathbf{H}_{X}$ onto a hypersphere via a nonlinear transformation $g_{\Phi}(\cdot)$ to obtain $\mathbf{Z}_X$. We use a normalized temperature-scaled cross-entropy loss (NT-Xent) \cite{wu2018unsupervised} with a temperature parameter $\rho$ to maximize mutual information between positive pairs, resulting in the \textit{\textbf{cell}ular complex \textbf{c}ontrastive \textbf{l}earning} (CellCL) loss:
\begin{equation}
    \mathcal{L}_{\mathrm{CellCL}} = -\frac{1}{N} \sum_{i=1}^{N} \log \frac{\exp(s(\mathbf{Z}_{X_i}, \tilde{\mathbf{Z}}_{X_i}) / \rho )}{\sum_{j=1}^{N} \mathbb{I}_{[j \neq i]} \exp(s(\mathbf{Z}_{X_i}, \tilde{\mathbf{Z}}_{X_j}) /  \rho )}, \label{loss_CellCL}
\end{equation}
here, $s(\mathbf{a}, \mathbf{b})$ denotes the cosine similarity between $\mathbf{a}$ and $\mathbf{b}$, defined as $s(\mathbf{a}, \mathbf{b}) = \frac{\mathbf{a}^\top \mathbf{b}}{|\mathbf{a}| |\mathbf{b}|}$. $\mathbb{I}_{[j \neq i]}$ is an indicator function that excludes other samples outside the positive pair, and $\rho$ is the temperature parameter that controls the sensitivity of contrastive learning.

\subsection{Adaptive Trimming of Cellular Topological Redundancy}
\label{redundant}
Through the framework introduced in Section \ref{contrastive}, we have obtained a complete set of high-order information while preserving the original graph skeleton. However, not all of this high-order information is necessarily semantically meaningful; in fact, some of it may even degrade the performance of downstream tasks. We refer to this key observation as \textbf{Cellular Topological Redundancy}. As illustrated in the motivation Figure \ref{fig:motivation}, when a fixed proportion of 2-cells is removed (e.g., 10\% interval in the figure), certain data points exhibit improved performance compared to the baseline with the complete set of 2-cells. This evidence suggests that some high-order topological information can actually degrade task-relevant information. To address the issue of cellular topological redundancy, a \textit{\textbf{Cell}ular \textbf{Trim}ming Scheduler} (CellTrim) adaptively removes 2-cells from the original cellular distribution $P_{\mathcal{C}}(X)$, constructing a refined distribution where high-order structures are task-relevant and semantically meaningful. This trimming scheduler is optimized by coupling cellular sparsification loss with contrastive learning task gradients through bi-level meta-learning.

\textbf{Cellular Trimming Scheduler $\Psi(\tau_{\alpha})$. }  For a given cellular complex $X$, the original distribution of 2-cells is denoted as $P_{\mathcal{C}}(X)\in \{0,1\}^{N_2}$, which includes the entire set of 2-cells $\mathcal{C}^{(2)}=\{\tau^{(2)}_{\alpha}\}_{\alpha=1}^{N_2}$, where $N_r$ represents the number of $r$-cells in $X$. Suppose that after $L$ layers of message passing, the embedding of a 2-cell $\tau^{(2)}_{\alpha}$ is given by $h^{(L)}_2(\tau_{\alpha})\in \mathbb{R}^d$.  We wish to decide whether to retain or trim the 2-cell by modeling a cellular trimming scheduler $\Psi$ via a categorical distribution. In particular, we assume that the decision variable, represented as a two-dimensional vector $\mathbf{y}=(y_{\alpha,0}, y_{\alpha,1})$, is sampled from a categorical distribution $\text{Cat}(\lambda_{\alpha})$, where the two entries correspond to “trim” (index 0) and “retain” (index 1), respectively. The probability vector $\lambda{_\alpha} \in \mathbb{R}^2$ is conditioned on the embedding $h^{(L)}_2(\tau_{\alpha})$ as follows:
\begin{equation}
    \lambda_{\alpha}=\text{Softmax}(W_{\Psi} \cdot h^{(L)}_2(\tau_{\alpha})+b),
\end{equation}
with $W_{\Psi} \in \mathbb{R}^{2 \times d}$ being a trainable weight matrix. Direct sampling from $\operatorname{Cat}(\lambda_\alpha)$ is non-differentiable; hence, to enable back-propagation, we adopt the Gumbel-Softmax trick \cite{jang2016categorical, maddison2016concrete}. Concretely, for each category $i \in \{0,1\}$ we compute
\begin{equation}
    y_{\alpha,i} = \frac{\exp \left( (\log \lambda_{\alpha,i} + g_i) / \zeta \right)}{\sum_{j=0}^{1} \exp \left( (\log \lambda_{\alpha,j} + g_j) / \zeta \right)},
\end{equation}
where $g_i \sim \operatorname{Gumbel}(0,1)$ is noise sampled from the Gumbel distribution and $\zeta > 0$ is a temperature parameter controlling the approximation’s smoothness. As $\zeta \to 0$, the vector $\mathbf{y}_\alpha = (y_{\alpha,0}, y_{\alpha,1})$ approaches a one-hot vector, we can directly use $y_{\alpha,1}$ as the indicator for retaining the 2-cell. Thus, we obtain the cellular trimming scheduler as $\Psi(\tau_{\alpha})=y_{\alpha,1}$, which indicates whether the embedding $h^{(L)}_2(\tau_{\alpha})$ contributes to the final embedding of the cellular complex. The overall embedding of the cellular complex is computed as
\begin{equation}
    \mathbf{H}_{X}^{'}=\sum_{r=0}^{1}\text{MLP}_{X,r}(\sum h^{(L)}_r(\tau))+\text{MLP}_{X,2}(\sum_{\alpha=1}^{N_2} \Psi(\tau_{\alpha}) \cdot h^{(L)}_2(\tau_{\alpha})). \label{trim}
\end{equation}

\textbf{Bi-level Meta-learning Optimization Process.} Recalling that the encoder of the cellular complex network is defined as $\mathbf{H}_X = f(X; \Theta)$, the parameter perturbation-based augmented embedding is represented as $\tilde{\mathbf{H}}_X = f(X; \tilde{\Theta})$. These embeddings are processed through a shared projection head, yielding $\mathbf{Z}_X = g(\mathbf{H}_X; \Phi)$ and $\tilde{\mathbf{Z}}_X = g(\tilde{\mathbf{H}}_X; \Phi)$. Let the cellular trimming scheduler be denoted as $\Psi(\tau_{\alpha}; \Upsilon)$. As formulated in Equation \eqref{trim}, it produces the refined embeddings $\mathbf{H}_{X}^{'}$ and $\tilde{\mathbf{H}}_X^{'}$. These embeddings are then projected to obtain $\mathbf{Z}_X^{'}$ and $\tilde{\mathbf{Z}}_X^{'}$, which are used for training with the contrastive loss in Equation \eqref{loss_CellCL}. Our objective is to jointly learn the parameters of the Cellular Trimming Scheduler $\Upsilon$ along with the contrastive network parameters $\{\Theta, \Phi\}$.

\begin{table*}[t]
	\begin{center}
		\begin{small}
			\begin{tabular}{c|cccc|cc|c}
				\hline
                \hline
				\text{Dataset} & \text{NCI1} & \text{PROTEINS} & \text{MUTAG} 
& \text{NCI109} & \text{IMDB-B} 
& \text{IMDB-M} & A.R. $\downarrow$ \\
				\hline
                    \hline
                    \text{node2vec}  & 54.9 $\pm$ 1.6 & 57.5 $\pm$ 3.6 & 72.6 $\pm$ 10.0 
& -& - 
& -& 11.0\\
                    \text{sub2vec}  & 52.8 $\pm$ 1.5 & 53.0 $\pm$ 5.6 & 61.1 $\pm$ 15.8 
& -& 55.3 $\pm$ 1.5 
& -& 11.8\\
                    \text{graph2vec}  & 73.2 $\pm$ 1.8 & 73.3 $\pm$ 2.0 & 83.2 $\pm$ 9.3 
& -& 71.1 $\pm$ 0.5 
& -& 9.3\\
                    \text{InfoGraph}  & 76.2 $\pm$ 1.0 & 74.4 $\pm$ 0.3 & \underline{89.0 $\pm$ 1.1} 
& 76.2 $\pm$ 1.3 & \underline{73.0 $\pm$ 0.9} 
&  48.1 $\pm$ 0.3 & 5.7\\
                    \text{GraphCL}  & 77.9 $\pm$ 0.4 & 74.4 $\pm$ 0.5 & 86.8 $\pm$ 1.3 
& 78.1 $\pm$ 0.4& 71.2 $\pm$ 0.4 
& 48.9 $\pm$ 0.3& 6.2\\
                    \text{ADGCL}  & 73.9 $\pm$ 0.8 & 73.3 $\pm$ 0.5 & 88.7 $\pm$ 1.9 
& 72.4 $\pm$ 0.4& 70.2 $\pm$ 0.7 
& 48.1 $\pm$ 0.4& 8.3\\
                    \text{JOAO}  & 78.1 $\pm$ 0.5 & 74.6 $\pm$ 0.4 & 87.4 $\pm$ 1.0 
& 77.2 $\pm$ 0.6 & 70.2 $\pm$ 3.1 
& 48.9 $\pm$ 1.2 & 6.7\\
                    \text{JOAOv2}  & 78.4 $\pm$ 0.5 & 74.1 $\pm$ 1.1 & 87.7 $\pm$ 0.8 
& 78.2 $\pm$ 0.8 & 70.8 $\pm$ 0.3 
& 49.2 $\pm$ 0.9 & 5.3\\
                    \text{RGCL}  & 78.1 $\pm$ 1.0 & 75.0 $\pm$ 0.4 & 87.7 $\pm$ 1.0 
& 77.7 $\pm$ 0.3& 71.9 $\pm$ 0.9 
& \underline{49.3 $\pm$ 0.4} & 4.2\\
                    \text{SimGRACE}  & \underline{79.1 $\pm$ 0.4} & \underline{75.3 $\pm$ 0.1} & \underline{89.0 $\pm$ 1.3} 
& \underline{78.4 $\pm$ 0.4} & 71.3 $\pm$ 0.8 & 49.1 $\pm$ 0.8 & \underline{2.7} \\
                    \text{HTML} & 78.2 $\pm$ 0.7 & 75.0 $\pm$ 0.3 & 88.9 $\pm$ 0.8 & 77.9 $\pm$ 0.2 & 71.7 $\pm$ 0.4 & 48.9 $\pm$ 0.6 & 4.0 \\

  \hline

 \text{\ours}& \textbf{79.4} $\pm$ 0.2 & \textbf{75.7 $\pm$ 0.1} & \textbf{89.7 $\pm$ 0.3} & \textbf{78.9 $\pm$ 0.4} & \textbf{73.4 $\pm$ 0.1} & \textbf{50.6 $\pm$ 0.2} & \textbf{1.0}\\
 \hline
 \hline
			\end{tabular}
		\end{small}
	\end{center}
 \caption{Unsupervised representation learning classification accuracy (\%) on TU datasets. A.R denotes the average rank of the results. The best results are highlighted in \textbf{bold}, and the second best results are highlighted with \underline{underline}.}
	\label{tab:cwn unsupervised learning}
\end{table*}

The overall training procedure consists of two stages. In the first conventional training stage, the cellular trimming scheduler $\Psi$ remains fixed while minimizing the Cellular Complex Contrastive Learning loss $\mathcal{L}_{\text{CellCL}}$, which is computed over the trimmed cellular distribution in a standard manner. In the second stage, a meta-learning-based approach is employed, wherein $\Psi$ is updated via a bi-level optimization process. The goal is to guide $\Psi$ towards suppressing the gradient contributions of higher-order 2-cells that contain task-irrelevant information in the contrastive learning task loss. To achieve this, the cellular sparsification loss is computed with respect to the performance of $f(\cdot; \Theta)$ and $g(\cdot; \Phi)$, which is measured using the gradients of $f(\cdot; \Theta)$ and $g(\cdot; \Phi)$ during the backpropagation of contrastive loss, formulated as follows:
\begin{equation}
\begin{gathered}
        \min_{\Upsilon } \mathcal{L}_{\mathrm{CellCL}}\left(\{\mathbf{Z}_X^{'},\tilde{\mathbf{Z}}_X^{'}\};\Theta ^{\star}(\Upsilon ), \Phi^{\star}(\Upsilon )\right), \\
        \text{where} \ \Theta ^{\star}(\Upsilon ), \Phi^{\star}(\Upsilon )= \underset{\Theta , \Phi }{\arg \min} \mathcal{L}_{\mathrm{CellCL}}\left(\{\mathbf{Z}_X^{'},\tilde{\mathbf{Z}}_X^{'}\};\Theta ,\Phi ,\Upsilon \right ). \label{bi-level}
\end{gathered}
\end{equation}

At iteration $k$, we adopt a second-derivative technique \cite{liu2019self, liu2018darts} to approximate $\Theta^{\star}(\Upsilon) \approx \hat{\Theta}^{(k+1)}(\Upsilon^{(k)})$ and $\Phi^{\star}(\Upsilon) \approx \hat{\Phi}^{(k+1)}(\Upsilon^{(k)})$ by performing a single gradient step from the current contrastive network parameters $\Theta^{(k)}$ and $\Phi^{(k)}$, as follows:
\begin{equation}
    \begin{gathered}
        \hat{\Theta}^{(k+1)}(\Upsilon^{(k)})=\Theta^{(k)}-\alpha \nabla_{\Theta^{(k)}}\mathcal{L}\left(\{\mathbf{Z}_X^{'},\tilde{\mathbf{Z}}_X^{'}\};\Theta^{(k)} ,\Phi^{(k)},\Upsilon^{(k)}\right), \\
        \hat{\Phi}^{(k+1)}(\Upsilon^{(k)})=\Phi^{(k)}-\beta \nabla_{\Phi^{(k)}}\mathcal{L}\left(\{\mathbf{Z}_X^{'},\tilde{\mathbf{Z}}_X^{'}\};\Theta^{(k)} ,\Phi^{(k)},\Upsilon^{(k)}\right),
    \end{gathered}
\end{equation}
where $\alpha$ and $\beta$ are the respective learning rates. Finally, the update of the cellular trimming scheduler $\Psi$ in Equation \eqref{bi-level} is transformed as follows:
\begin{equation}
    \Upsilon^{(k+1)}=\underset{\Upsilon^{(k)}}{\arg \min} \mathcal{L}_{\mathrm{CellCL}}\left(\{\mathbf{Z}_X^{'},\tilde{\mathbf{Z}}_X^{'}\};\hat{\Theta}^{(k+1)}(\Upsilon^{(k)}),\hat{\Phi}^{(k+1)}(\Upsilon^{(k)})\right).
\end{equation}

The training of $f(\cdot; \Theta)$ and $g(\cdot; \Phi)$ alternates with the bi-level optimization process for training $\Psi(\cdot; \Upsilon)$ until convergence.

\begin{table*}[t]
	\begin{center}
	\begin{small}
			\begin{tabular}{c|ccc|cc|c}
				\hline
                \hline
				\text{Dataset} & \text{NCI1} & \text{PROTEINS}
& \text{NCI109} & \text{IMDB-B} 
& \text{IMDB-M} & A.R. $\downarrow$ \\
				\hline
                    \hline
                    \text{InfoGraph}  & 68.9 $\pm$ 0.8 & 69.5 $\pm$ 0.9    
& 67.8 $\pm$ 0.7 & 68.5 $\pm$ 0.6 
&  43.8 $\pm$ 0.2 & 6.4\\
                    \text{GraphCL}  & 69.6 $\pm$ 0.3 & 69.7 $\pm$ 0.1   
& \underline{69.2 $\pm$ 0.3} & 67.8 $\pm$ 1.3 
& 43.9 $\pm$ 0.3 & 4.4\\
                    \text{ADGCL}  & 65.6 $\pm$ 0.6 & 68.5 $\pm$ 2.0 
& 65.7 $\pm$ 0.8 & 67.7 $\pm$ 0.9 
& 41.9 $\pm$ 0.9& 8.8\\
                    \text{JOAO}  & 69.1 $\pm$ 0.1 & 70.9 $\pm$ 2.0  
& \underline{69.2 $\pm$ 0.3} & 67.6 $\pm$ 0.8 
& 43.9 $\pm$ 1.0 & 5.0\\
                    \text{JOAOv2}  & 69.1 $\pm$ 0.4 & 70.1 $\pm$ 1.4  
& \underline{69.2 $\pm$ 0.2} & 68.3 $\pm$ 1.0 
& 43.6 $\pm$ 1.1 & 5.2\\
                    \text{RGCL}  & \underline{70.2 $\pm$ 0.7} & 71.2 $\pm$ 0.9 
& 69.1 $\pm$ 0.3 & \textbf{68.9 $\pm$ 0.3} 
& \textbf{44.5 $\pm$ 0.9} & \underline{2.4} \\
                    \text{SimGRACE}  & 69.3 $\pm$ 0.3 & \underline{71.3 $\pm$ 0.7}  
& 69.0 $\pm$ 0.2 & \underline{68.6 $\pm$ 0.7} & \underline{44.2 $\pm$ 0.6} & 3.4\\
                    \text{HTML} & 69.4 $\pm$ 0.1 & 69.1 $\pm$ 1.5 & 68.9 $\pm$ 0.1 & 68.4 $\pm$ 0.3 & 43.2 $\pm$ 0.3 & 6.4\\
        
  \hline
 \text{\ours}& \textbf{70.4 $\pm$ 0.5} & \textbf{71.8 $\pm$ 0.1}  & \textbf{69.5 $\pm$ 0.3} & 68.5 $\pm$ 0.4 & 44.0 $\pm$ 0.3  & \textbf{1.8}\\
 \hline
 \hline
			\end{tabular}
			\end{small}
	\end{center}
 \caption{Semi-supervised representation learning classification accuracy (\%) on TU datasets.}
	\label{tab:cwn semisupervised learning}
\end{table*}

\section{Theoretical Justification of CellCLAT}
\label{theoretical}
\textbf{Topological Expressiveness of CellCLAT.} We characterize the discriminability of CellCLAT's encoder: CCNN, in comparison to GNNs by analyzing its relationship with the WL test. The following theorem provides a rigorous proof that the topological expressiveness of CCNN is strictly stronger than that of the WL test, demonstrating its superior ability to recognize higher-order structures.
\begin{theorem}[CCNN is strictly more expressive than the WL test] \label{expressive}
    Let $f:\mathcal{G} \to \mathcal{X}$ be a skeleton‐preserving gluing process: from graphs to cellular complexes. Let $G_1, G_2$ be graphs such that the 1‐WL test (and hence any GNN that is bounded by WL) cannot distinguish between them, i.e., $\mathrm{c}^{G_1,t} = \mathrm{c}^{G_2,t}$ for all iterations $t$. Then, there exists an iteration $t^\star$ such that the $\text{CCNN}$ colouring $\mathrm{b}^{f(G),t^\star}$ of the 0‐cells of the lifted complexes $f(G_1)$ and $f(G_2)$ satisfies $\mathrm{b}^{f(G_1),t^\star} \neq \mathrm{b}^{f(G_2),t^\star}$, implying that $\text{CCNN}$ distinguishes $G_1$ and $G_2$.
\end{theorem}

We adopt the techniques from \cite{bodnar2021weisfeiler2}, with notation defined as $\mathrm{c}^{G,t}$ and $\mathrm{b}^{f(G),t^\star}$, and the detailed proof is provided in Appendix \ref{proof1}.

\textbf{Causal Analysis of Cellular Topological Redundancy.} We employ causal inference techniques \cite{pearl2009causality, pearl2016causal} to theoretically analyze the motivational experimental phenomena (Figure \ref{fig:motivation}). By constructing a bottom-up structural causal model (SCM) \cite{pearl2016causal} that underlies the participation of high-order cellular topology in the message-passing process, we demonstrate that redundant cellular topology acts as a \textbf{confounder}. 

We define the causal variables as follows: $E$ represents the graph-level embedding, $\hat{Y}$ denotes the predicted label, and $T$ corresponds to the high-order cellular topology within the cell complex $X$ constructed over the graph $G$, i.e., the distribution of 2-cells $P_{\mathcal{C}}(X)$. The corresponding SCM graph is illustrated in Figure \ref{fig:scm}. We explain the relationships of each dependence edge in SCM: 
\begin{itemize}
    \item $E \rightarrow \hat{Y}$. This dependence arises from the forward computation of $\text{CCNN}$, where the graph-level embedding computed under the learned parameters of the current epoch is used for label prediction.
    \item $T \rightarrow E$. The embedding update is influenced by the aggregation of 2-cells during the message-passing process.
    \item $T \rightarrow \hat{Y}$. The influence of $T$ on the downstream task can be either beneficial or detrimental. This corresponds to the motivational experiment where randomly removing 2-cells (altering their distribution) affects performance, as observed in the scatter plots of the baseline.
\end{itemize}
The backdoor path $E \leftarrow T \rightarrow \hat{Y}$ \cite{pearl2009causality} prevents us from learning a stable causal relationship $E \rightarrow \hat{Y}$, thereby proving that redundant cellular topology serves as a confounder. To robustly identify the causal effect of $E$ on $\hat{Y}$, we adopt the $do$-operation from causal inference. Formally, we train the model by optimizing $P(\hat{Y} \mid do(E))$: 
\begin{equation}
    P(\hat{Y} \mid do(E)) = \int P(\hat{Y} \mid E, T) P(T) \, dT.\label{backdoor}
\end{equation}

Each value of $T = t_i$ can be estimated using the cellular trimming scheduler $\Psi$. The scheduler $\Psi$ dynamically trims 2-cells to fit the new distribution $T = t_i$. The term $P(\hat{Y} \mid E, T=T_i)$ represents the prediction obtained from the embedding after trimming $T$. Thus, the implementation of CellCLAT to address cellular topological redundancy can be interpreted through the backdoor adjustment formula (Equation \eqref{backdoor}) applied to confounder $T$. For further details on the causal background, please refer to the Appendix \ref{causal}.

\section{Experiments}
\subsection{Experimental Setup}

\begin{table*}[t]
	\begin{center}
	\begin{small}
			\begin{tabular}{c|cccc|cc}
				\hline
                \hline
				\text{Dataset} & \text{NCI1} & \text{PROTEINS} & \text{MUTAG} 
& \text{NCI109} & \text{IMDB-B} 
& \text{IMDB-M} \\
				\hline
                    \hline
                    \text{SimGRACE}  & 79.1 $\pm$ 0.4 & 75.3 $\pm$ 0.1 & 89.0 $\pm$ 1.3 
& 78.4 $\pm$ 0.4 & 71.3 $\pm$ 0.8 & 49.1 $\pm$ 0.8 \\
                    \text{CellCL}& 79.3 $\pm$ 0.3 & 75.6 $\pm$ 0.1 & 89.4 $\pm$ 0.3 & 78.6 $\pm$ 0.3 & 73.0 $\pm$ 0.4 & 50.3 $\pm$ 0.1\\
                    \text{CellCL w 0-CellTrim} & 78.7 $\pm$ 0.2 & 75.4 $\pm$ 0.2 & 88.9 $\pm$ 0.8 & 77.8 $\pm$ 0.1 & \textbf{73.5 $\pm$ 0.1} & 50.5 $\pm$ 0.3 \\
                    \text{CellCL w 1-CellTrim} & 79.3 $\pm$ 0.2 & 75.6 $\pm$ 0.2 & 88.5 $\pm$ 0.9 & 78.2 $\pm$ 0.1 & \textbf{73.5 $\pm$ 0.2} & 50.4 $\pm$ 0.3 \\
 \text{\ours\ (CellCL w 2-CellTrim)}& \textbf{79.4 $\pm$ 0.2} & \textbf{75.7 $\pm$ 0.1} & \textbf{89.7 $\pm$ 0.3} & \textbf{78.9 $\pm$ 0.4} & 73.4 $\pm$ 0.1 & \textbf{50.6 $\pm$ 0.2} \\
 \hline
 \hline
			\end{tabular}
			\end{small}
	\end{center}
 \caption{Ablation studies on the unsupervised settings.}
	\label{tab:ablation study}
\end{table*}

\textbf{Datasets.}
For unsupervised learning, we benchmark our proposed \ours\ on six established datasets in TU datasets \cite{tudataset} including four bioinformatics datasets (NCI1, PROTEINS, MUTAG, NCI109) and two social network networks datasets (IMDB-B, IMDB-M). For semi-supervised learning, we use the same datasets as in the unsupervised setting, except for MUTAG, as its small size leads to severe class imbalance issues during k-fold cross-validation in downstream tasks. 

\textbf{Compared Baselines.}
For unsupervised learning, we compare \ours\ with ten unsupervised baselines including Node2Vec \cite{node2vec}, Sub2Vec \cite{sub2vec}, Graph2Vec \cite{graph2vec}, InfoGraph \cite{infograph}, GraphCL \cite{graphcl}, ADGCL \cite{adgcl}, JOAO \cite{joao}, RGCL \cite{rgcl}, SimGRACE \cite{simgrace} and HTML \cite{html}. In the semi-supervised setting, we select seven unsupervised baselines for comparison.

\textbf{Evaluate Protocols.}
For unsupervised learning, we employ the complete datasets to train \ours\  on unsupervised datasets. Subsequently, these representations are utilized as input to a downstream SVM classifier with 10-fold cross-validation. We run five times with different seeds on each dataset and the mean and standard deviation of classification accuracy is reported.
For semi-supervised learning, the training phase is the same as the unsupervised setting. In the downstream classification task, we utilize only 10\% of the labeled data to train a SVM classifier, while leveraging the remaining unlabeled data through pseudo-labeling. Specifically, we randomly select 10\% of the training set as labeled data and train an initial classifier. The classifier is then used to generate pseudo-labels for the remaining unlabeled data, and a final model is trained using both the labeled and pseudo-labeled data.
To ensure reliability, we run each experiment with five random seeds and report the mean and standard deviation of accuracy. The code is available at: \url{https://github.com/ByronJi/CellCLAT}.

\subsection{Unsupervised Learning}
The results of unsupervised graph-level representation learning for downstream graph classification tasks are presented in Table \ref{tab:cwn unsupervised learning}. Our method consistently achieves the best performance across multiple datasets, obtaining the lowest average rank 1.0 among all methods, highlighting the effectiveness of our approach in capturing discriminative representations. In addition, our method has smaller variance compared to other contrastive learning methods, which further demonstrates the stability of our method.

\subsection{Semi-supervised Learning}
Table \ref{tab:cwn semisupervised learning} presents the results of semi-supervised graph-level representation learning for downstream classification tasks. Our method, CellCLAT, achieves the lowest average rank 1.8 among all compared methods, demonstrating the best overall performance across multiple datasets. This result indicates that the representations learned during pretraining are highly effective, enabling strong performance even when labeled data is sparse in downstream tasks.

\begin{figure}
    \centering
    \includegraphics[width=1.0\linewidth]{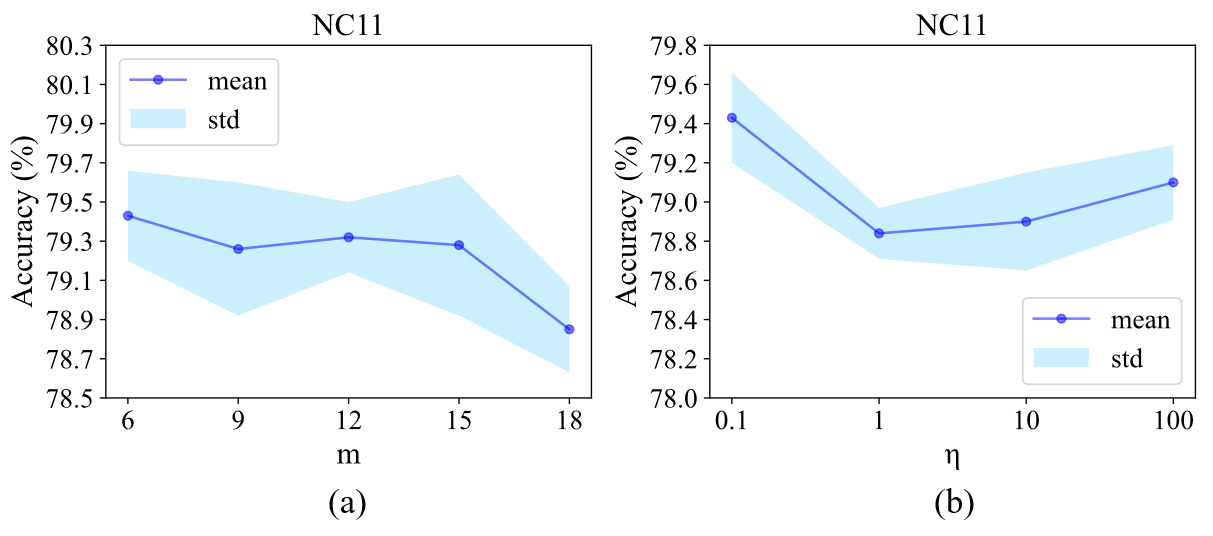}
    \caption{Hyper-parameter sensitivity analysis.}
    \label{fig:hyper-m-k}
\end{figure}

\subsection{Ablation Studies}
We conduct ablation studies on the unsupervised settings, shown in Table \ref{tab:ablation study}. Simgrace is a graph-based contrastive learning method without data augmentation, whereas CellCL, as described in Section \ref{contrastive}, employs cellular complex contrastive learning to obtain \textbf{all complete representations} of 0-cells, 1-cells, and 2-cells. While our motivation experiments demonstrate semantic redundancy in 2-cells, we aim to investigate whether lower-order cells exhibit similar phenomena. To this end, we slightly modify the notation of CellTrim: 0-CellTrim, 1-CellTrim, and 2-CellTrim denote the application of the cellular trimming scheduler to 0-cells (nodes), 1-cells (edges), and 2-cells (polygons), respectively. Interestingly, trimming 0-cell representations (CellCL w/ 0-CellTrim) and 1-cell representations (CellCL w/ 1-CellTrim) leads to performance \textbf{degradation} to varying degrees compared to using the complete cellular representation (CellCL). Notably, 0-cells demonstrate greater significance, as trimming 0-cells leads to the most substantial performance degradation. As anticipated, trimming 2-cell representations yields the most significant improvement, validating that CellCLAT effectively trims redundant topological information.

\subsection{Hyper-parameter Sensitivity Analysis}

\textbf{Ring size.}
The ring size refers to an upper bound $m$ on the number of edges in polygons or rings during the gluing process. Figure \ref{fig:hyper-m-k}(a) presents the classification accuracy on the NCI1 dataset for different values of $m$, ranging from 6 to 18. We observe that on the NCI1 dataset, the highest accuracy is achieved when $m=6$, and as $m$ increases, the performance gradually declines. This may be because NCI1 primarily consists of functional groups that contain six-membered rings (e.g., benzene), making $m=6$ the most effective choice in our method.

\textbf{Permuted rate.}
We analyze the effect of the permuted rate $\eta$ on model performance using the NCI1 dataset. As shown in Figure \ref{fig:hyper-m-k}(b), the classification accuracy is highest when $\eta=0.1$.

\begin{table}[t]
    \setlength{\tabcolsep}{4pt}
    \begin{center}
    \begin{small}
            \begin{tabular}{c|cccccc}
                \hline
                \hline
                \text{Dataset} & \text{NCI1} & \text{PROTEINS} & \text{MUTAG} 
                & \text{NCI109} & \text{IMDB-B} & \text{IMDB-M} \\
                \hline
                \hline
                \text{Time} & 15s & 14s & 1s & 15s & 37s & 54s \\
                \hline
                \hline
            \end{tabular}
            \end{small}
    \end{center}
    \caption{Gluing time across different datasets.}
    \label{tab:gluing-time}
\end{table}

\begin{table}[t]
    \begin{center}
    \begin{small}
            \begin{tabular}{c|cccc}
                \hline
                \hline
                \text{Dataset} & \text{GraphCL} & \text{RGCL} & \text{HTML} & \text{Ours} \\
                \hline
                \hline
                \text{PROTEINS} & 17s & 41s & 43s & 51s \\
                \text{NCI1} & 94s & 134s & 148s & 172s \\
                \hline
                \hline
            \end{tabular}
            \end{small}
    \end{center}
    \caption{Training time across different methods.}
    \vspace{-0.35cm}
    \label{tab:time-comparison}
\end{table}

\subsection{Complexity and Efficiency Analysis}
The complexity of CellCLAT can be divided into two components: 1) the gluing process lifting a graph to a cellular complex, and 2) the message passing procedure.

First, gluing process is a one-time preprocessing step before training, executable in $\Theta((|E|+|V|N)\text{polylog}|V|)$ time \cite{ferreira2014amortized}, where $N$ is the number of induced cycles (upper bounded by a small, dataset-dependent constant). We report the gluing times for TU datasets in Table \ref{tab:gluing-time}, showing that it scales approximately linearly with the size of the input graph.

Second, for an $r$-cell $\sigma$ in a cellular complex $X^{(2)}$ with maximum boundary size $B_r$, the computational complexity of the four types of messages is given by
\begin{equation}
    \Theta\left(\sum_{r=0}^{2} B_r N_r + \binom{B_r}{2} N_r + B_{r+1} N_{r+1} + \binom{B_{r+1}}{2} N_{r+1}\right),
\end{equation}
simplified as $\Theta\left(\sum_{r=0}^{2} \binom{B_{r+1}}{2} N_r\right)$, where $N_r$ denotes the number of $r$-cells. Standard GNNs are a special case ($r \in {0,1}$) with $\Theta(N_0 + N_1) = \Theta(|V| + |E|)$. Therefore, our model's runtime (include bi-level meta cellular trimming process) is comparable to GNNs, with training times in Table \ref{tab:time-comparison}.

\subsection{Visualization Results}
\begin{figure}
    \centering
    \includegraphics[width=1.0\linewidth]{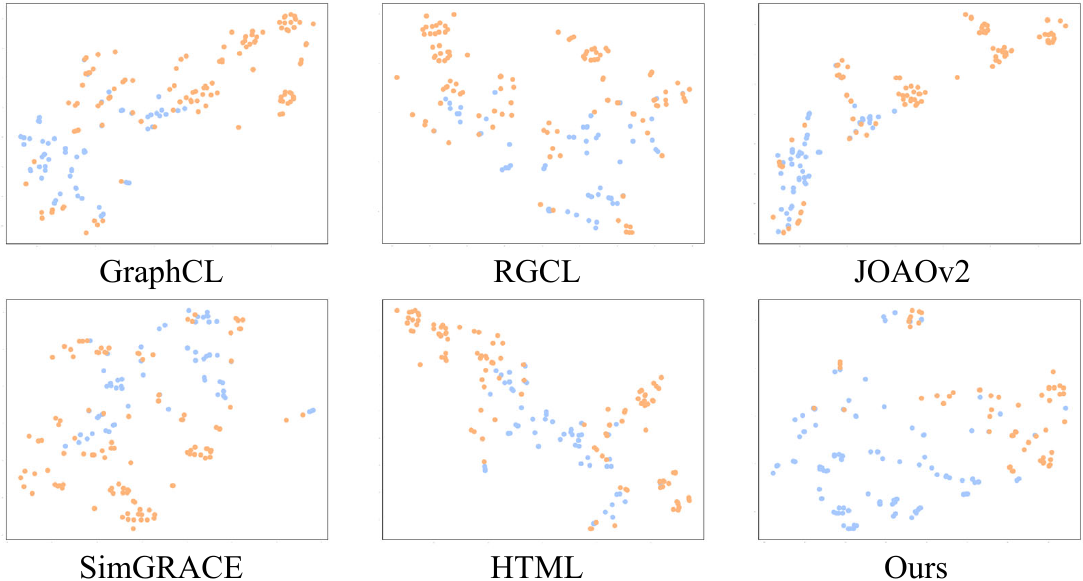}
    \caption{t-SNE visualization of six methods on MUTAG.}
    \label{fig:tsne}
\end{figure}

In Figure \ref{fig:tsne}, we present the t-SNE visualization of six different methods on the MUTAG dataset. Our method exhibits a more distinct and well-separated clustering structure, indicating its ability to learn discriminative representations compared to the baselines. 

\section{Conclusion}
In this paper, we are the first to propose a self-supervised framework for learning cellular complex representations, namely CellCLAT, which overcomes the challenge of extrinsic structural constraints inherent to cellular complex spaces. Furthermore, we unveil a key phenomenon: Cellular Topological Redundancy, which introduces significant performance degeneration for understanding the higher-order topological structures on downstream tasks. Our approach offers the first self-supervised method capable of learning semantic-rich representations from cellular complexes. The promising results of our framework highlight its potential for more expressive modeling of complex data relations and encourage further exploration of SSL techniques in combinatorial topological spaces.

\begin{acks}
    The authors would like to thank anonymous reviewers for their valuable comments. This work is supported by National Natural Science Foundation of China, Grant No. 62406313, Postdoctoral Fellowship Program, Grant No. GZC20232812, China Postdoctoral Science Foundation, Grant No. 2024M753356, 2023 Special Research Assistant Grant Project of the Chinese Academy of Sciences, 2023 BDA project (MIIT), Basic Research Project of the Institute of Software, Chinese Academy of Sciences (Project No. ISCAS-JCZD-202402) and National Key R\&D Program of China (No. 2023YFB3002901).
\end{acks}


\bibliographystyle{ACM-Reference-Format}
\balance
\bibliography{sample-base}

\appendix
\section{Derivation of Theoretical Justification}
\subsection{Proof of Theorem 1}
\label{proof1}
\begin{definition}[Color Refinement on Cellular Complexes] \label{def1}
Let $X$ be a cellular complex, and let the set of cells in $X$ be denoted as $\mathcal{C}(X)$. We define a color mapping $\mathrm{c}:\mathcal{C}(X)\rightarrow \mathbb{N}$, which assigns cells to a set of natural numbers (i.e., colors). The iterative process of color refinement is represented as follows: at iteration $t$, the color of a cell $\sigma$ is denoted by $\mathrm{c}_{\sigma}^{X,t}$. The process continues until some iteration $t^{\star}$ satisfies $\mathrm{c}_{\sigma}^{X,t^{\star}}=\mathrm{c}_{\sigma}^{X,t^{\star}+1}$, at which point we say the color mapping $\mathrm{c}$ has converged. The final color of any cell $\sigma \in X$ is denoted as $\mathrm{c}_{\sigma}^{X,\infty}$. Specifically, for a given cell $\sigma \in X$, we denote its final color as $\mathrm{c}_{\sigma}^{X}$ or $\mathrm{c}^{X}(\sigma)$.
\end{definition}

\begin{definition}[Color Equivalence on Cellular Complexes] \label{def2}
    Let $X_1$ and $X_2$ be two distinct cellular complexes, and let $\mathrm{c}$ be a cellular color refinement. After applying color refinement $\mathrm{c}$, the color sets of $X_1$ and $X_2$ are given by $\mathrm{c}^{X_1}=\{\!\!\{\mathrm{c}_{\sigma}^{X_1}|\forall \sigma \in X_1 \}\!\!\}$ and $\mathrm{c}^{X_2}=\{\!\!\{\mathrm{c}_{\tau}^{X_2}|\forall \tau \in X_2 \}\!\!\}$, respectively. We say that $X_1$ and $X_2$ are color-equivalent under $\mathrm{c}$, denoted as $X_1\sim_{\mathrm{c}}X_2$, if and only if $\mathrm{c}^{X_1}=\mathrm{c}^{X_2}$.
\end{definition}

In a more intuitive sense, this definition implies that the number of cells with dimension $n$ and a given color in $X_1$ equals the number of cells with the same color and dimension $n$ in $X_2$. Color equivalence reflects some structural similarities between the two complexes but does not imply isomorphism. This depends on the specific Color Refinement used.

Next, we define a measure of the ability of different Color Refinement (CR) methods to distinguish non-isomorphic graphs, allowing for comparison of the topological expressiveness of two CR models.

\begin{definition}[Topological Reduction] \label{def3}
    Let $\mathrm{a}$ and $\mathrm{b}$ be two CR models. We say that $\mathrm{a}$ can be reduced to $\mathrm{b}$, denoted as $\mathrm{a}\preceq \mathrm{b}$, if for all cellular complexes $X_1$ and $X_2$, and for all cells $\sigma \in X_1$ and $\tau \in X_2$ with dim($\sigma$)=dim($\tau$), the following condition holds:
\begin{equation}
    \label{reduce}
\mathrm{b}_{\sigma}^{X_1}=\mathrm{b}_{\tau}^{X_2}\Longrightarrow\mathrm{a}_{\sigma}^{X_1}=\mathrm{a}_{\tau}^{X_2}
\end{equation}
\end{definition}

Topological reduction, combined with Definition \ref{def2}, provides a means of comparing the topological expressiveness of two CR models. The equation $\mathrm{b}_{\sigma}^{X_1}=\mathrm{b}_{\tau}^{X_2}$ indicates that the CR model $\mathrm{b}$ cannot distinguish between non-isomorphic $X_1$ and $X_2$, so Equation \eqref{reduce} implies that if CR model $\mathrm{b}$ fails to distinguish non-isomorphic $X_1$ and $X_2$, then CR model $\mathrm{a}$ will also fail to do so. Therefore, if $\mathrm{a}\preceq \mathrm{b}$ holds, CR model $\mathrm{b}$ has stronger expressiveness in distinguishing non-isomorphic topological structures.

\begin{figure}[ht]
    \centering
    \includegraphics[width=0.5\linewidth,alt={Scm.}]{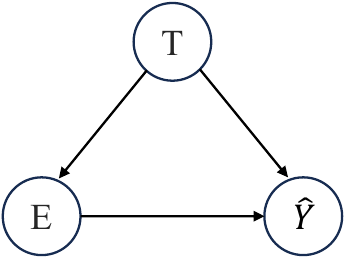}
    \caption{The proposed SCM graph for the CellCLAT framework. Redundant cellular topology $T$ acts as a \textbf{confounder}.}
    \label{fig:scm}
\end{figure}

\begin{figure}[ht]
   \centering
    \includegraphics[width=0.8\linewidth,alt={WL.}]{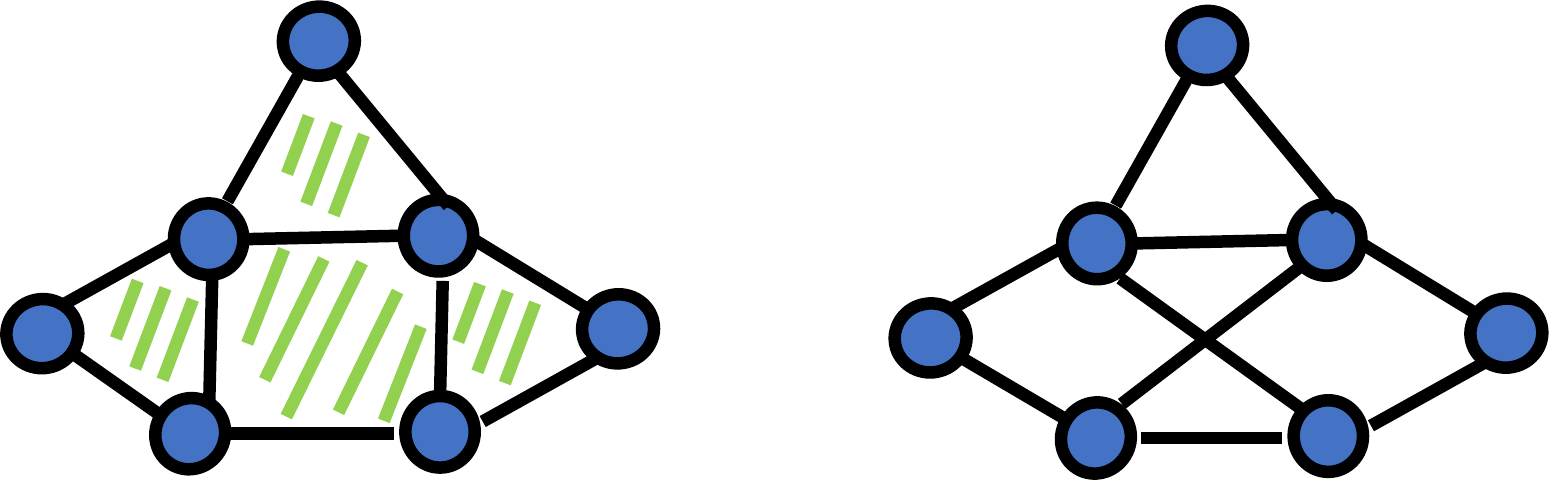}
    \caption{Two non-isomorphic graphs that cannot be distinguished by the WL test can be differentiated by CCNN, as the former contains triangular and quadrilateral 2-cells.}
    \label{fig:wl}
\end{figure}

With the aforementioned notations at hand, we proceed to prove the statement in Theorem \ref{expressive}: there exist non-isomorphic graphs $G_1$ and $G_2$ that cannot be distinguished by the 1-WL test (i.e., $\mathrm{c}^{G_1} = \mathrm{c}^{G_2}$), but can be differentiated by the $\text{CCNN}$ model (i.e., $\mathrm{b}^{f(G_1)} \neq \mathrm{b}^{f(G_2)}$). This implies that $\mathrm{b}^t \npreceq \mathrm{c}^t$. To do so, it suffices to find a pair of non-isomorphic graphs that cannot be distinguished by the 1-WL test but can be differentiated by the $\text{CCNN}$ network, as illustrated in Figure \ref{fig:wl}. To demonstrate that the topological expressiveness of $\text{CCNN}$ is strictly stronger than that of the 1-WL test, we also need to prove that $\mathrm{c}^t\preceq \mathrm{b}^t$. 

We firstly extend the color reduction for individual cells in Definition \ref{def3} to a color reduction for sets of cells.

\begin{lemma}\label{lemma1}
    Let $X_1$ and $X_2$ be two distinct cellular complexes, and let $A \subset X_1$ and $B \subset X_2$ be sets of cells, both of dimension $n$. For two CR models $\mathrm{a}$ and $\mathrm{b}$, if $\mathrm{a}\preceq \mathrm{b}$ and $\{\!\!\{\mathrm{a}_{\sigma}^{X_1}|\forall \sigma \in A \}\!\!\} \neq \{\!\!\{\mathrm{a}_{\tau}^{X_2}|\forall \tau \in B \}\!\!\}$, then $\{\!\!\{\mathrm{b}_{\sigma}^{X_1}|\forall \sigma \in A \}\!\!\} \neq \{\!\!\{\mathrm{b}_{\tau}^{X_2}|\forall \tau \in B \}\!\!\}$.
\end{lemma}

\textbf{Proof:} Assume the multiset $\{\!\!\{\mathrm{a}_{\sigma}^{X_1}|\forall \sigma \in A \}\!\!\}$ contains $k$ distinct colors $N_1,N_2,\dots,N_k$, where $N_k \in \mathbb{N}$ represents a specific color. Since $\{\!\!\{\mathrm{a}_{\sigma}^{X_1}|\forall \sigma \in A \}\!\!\} \neq \{\!\!\{\mathrm{a}_{\tau}^{X_2}|\forall \tau \in B \}\!\!\}$, there must exist a color $N_{\bullet}$ that appears with different frequencies in $\{\!\!\{\mathrm{a}_{\sigma}^{X_1}|\forall \sigma \in A \}\!\!\}$ and $\{\!\!\{\mathrm{a}_{\tau}^{X_2}|\forall \tau \in B \}\!\!\}$. Define $A^{\prime}=\left\{ \sigma \in A|\mathrm{a}_{\sigma}^{X_1}=N_{\bullet } \right\}$ and  \sloppy 
 $B^{\prime}=\left\{ \tau \in B|\mathrm{a}_{\tau}^{X_2}=N_{\bullet } \right\}$, then $\left| A^{\prime} \right| \neq \left| B^{\prime} \right|$.

Now, consider the cells $\gamma$ in $(A \cup B)\setminus(A^{\prime}\cup B^{\prime})$. Since $\mathrm{a}\preceq \mathrm{b}$, the converse-negative proposition of \eqref{reduce} implies that $\mathrm{a}_{\sigma}^{X_1}\neq\mathrm{a}_{\gamma}^{X_2} \Rightarrow \mathrm{b}_{\sigma}^{X_1}\neq\mathrm{b}_{\gamma}^{X_2}$. This means that the colors assigned by CR model $\mathrm{b}$ to the cells in $A^{\prime}\cup B^{\prime}$ differ from those in $(A \cup B)\setminus(A^{\prime}\cup B^{\prime})$.

By contradiction, assume $\{\!\!\{\mathrm{b}_{\sigma}^{X_1}|\forall \sigma \in A \}\!\!\} = \{\!\!\{\mathrm{b}_{\tau}^{X_2}|\forall \tau \in B \}\!\!\}$. Then, we would have $\{\!\!\{\mathrm{b}_{\sigma}^{X_1}|\forall \sigma \in A \setminus A^{\prime} \}\!\!\} = \{\!\!\{\mathrm{b}_{\tau}^{X_2}|\forall \tau \in B \setminus B^{\prime} \}\!\!\}$ and $\{\!\!\{\mathrm{b}_{\sigma}^{X_1}|\forall \sigma \in A^{\prime} \}\!\!\} = \{\!\!\{\mathrm{b}_{\tau}^{X_2}|\forall \tau \in B^{\prime} \}\!\!\}$. This leads to $\left| A^{\prime} \right|= \left| B^{\prime} \right|$, which is a contradiction. \hfill $\square$

\begin{theorem}
\label{theorem1}
Let $X_1$ and $X_2$ be two distinct cellular complexes, and let two CR models $\mathrm{a}$ and $\mathrm{b}$ satisfy $\mathrm{a}\preceq \mathrm{b}$. If $\mathrm{a}^{X_1}\neq \mathrm{a}^{X_2}$, then $\mathrm{b}^{X_1}\neq \mathrm{b}^{X_2}$.
\end{theorem}

\textbf{Proof:} By Lemma 1, when we traverse the sets of cells of different dimensions in $X_1$ and $X_2$, we obtain that if $\mathrm{a}^{X_1}=\{\!\!\{\mathrm{a}_{\sigma}^{X_1}|\forall \sigma \in X_1 \}\!\!\} \neq \{\!\!\{\mathrm{a}_{\tau}^{X_2}|\forall \tau \in X_2 \}\!\!\}=\mathrm{a}^{X_2}$, then $\mathrm{b}^{X_1}=\{\!\!\{\mathrm{b}_{\sigma}^{X_1}|\forall \sigma \in X_1 \}\!\!\} \neq \{\!\!\{\mathrm{b}_{\tau}^{X_2}|\forall \tau \in X_2 \}\!\!\}=\mathrm{b}^{X_2}$. \hfill $\square$

Theorem \ref{theorem1} states that for a pair of non-isomorphic cellular complexes $(X_1, X_2)$, $\mathrm{a}\preceq \mathrm{b}$ indicates that if model $\mathrm{a}$ can distinguish the non-isomorphic pair, then model $\mathrm{b}$ can as well. The aforementioned topological reduction theory provides tools for comparing the topological expressiveness of two CR models. $\mathrm{a}\preceq \mathrm{b}$ holds if and only if model $\mathrm{b}$ possesses at least the same topological expressiveness as model $\mathrm{a}$. Specifically, $\mathrm{a}\equiv \mathrm{b}$ holds if and only if the topological expressiveness of models $\mathrm{a}$ and $\mathrm{b}$ is identical.

According to the technical approach outlined above, we will now proceed to finally prove Theorem \ref{expressive}.

\textbf{Proof:} Consider the skeleton‐preserving gluing process $f:\mathcal{G} \to \mathcal{X}$, which maps any graph $G\in\mathcal{G}$ to a cellular complex $X=f(G)$ so that the vertices of $G$ correspond to the 0‐cells of $X$. Let 
\begin{equation}
    g_G : V_G \to P_{f(G)^{(0)}},
\end{equation}
be the isomorphism between the vertices of $G$ and the 0‐cells of $X$. For each graph $G$, let $\mathrm{c}^{G,t}$ denote the $\text{WL}$ colouring of $G$ at iteration $t$ and let $\mathrm{a}^{f(G),t}$ be the $\text{WL}$ colouring of the 1‐skeleton $f(G)^{(1)}$ induced by $g_G$ (i.e., for each $v\in V_G$, set $\mathrm{a}^{f(G)^{(1)},t}_{g(v)}:= \mathrm{c}^{G,t}_v$). Since the 1‐skeleton $f(G)^{(1)}$ is isomorphic to $G$, we have 
\begin{equation}
    \mathrm{a}^{f(G)^{(1)},t}= \mathrm{c}^{G,t}. \label{color}
\end{equation}

Let $\mathrm{b}^t$ denote the colouring (or feature labelling) of the 0‐cells as produced by the $\text{CCNN}$ at iteration $t$. By design, the $\text{CCNN}$ update for a cell $\tau$ is given by
\begin{equation}
        h^{(l+1)}_r(\tau)=\phi \left ( h^{(l)}_r(\tau), \bigotimes_{\mathcal{N}_i \in \mathcal{N}} \bigoplus_{\sigma \in \mathcal{N}_i(\tau)} \psi_{\mathcal{N}_i} \left( h^{(l)}_r(\tau), h^{(l)}_{r'}(\sigma), \Theta_i^{(l)} \right) \right ), 
\end{equation}
where the neighbourhoods are defined as follows: 
\begin{itemize}
    \item Boundary Adjacent: $\mathcal{B}(\tau)=\left\{ \sigma \mid \sigma \prec \tau \right\}$,
    \item Co-Boundary Adjacent: $\mathcal{C}(\tau)=\left\{ \sigma \mid \tau \prec \sigma \right\}$,
    \item Lower Adjacent: $\mathcal{N}_{\downarrow}(\tau)=\left\{ \sigma \mid \exists \ \delta \ \text{s.t.} \  \delta \prec \sigma \wedge \delta \prec \tau\right\}$,
    \item Upper Adjacent: $\mathcal{N}_{\uparrow}(\tau)=\left\{ \sigma \mid \exists \ \delta \ \text{s.t.} \ \sigma \prec \delta\wedge \tau \prec \delta \right\}$.
\end{itemize}

For all $G_1,G_2 \in \mathcal{G}$, our aim is to show that for all cellular complexes $X=f(G_1),Y=f(G_2)\in f(\mathcal{G})$ the $\text{WL}$ colouring $\mathrm{a}^t$ is topological reduced to the $\text{CCNN}$ colouring $\mathrm{b}^t$; that is, $\mathrm{a}^t\preceq \mathrm{b}^t$. By Theorem \ref{theorem1}, $\mathrm{a}^t\preceq \mathrm{b}^t$ signifies that if $\mathrm{a}^{f(G_1)^{(1)},t} \neq \mathrm{a}^{f(G_2)^{(1)},t}$, then it follows that $\mathrm{b}^{f(G_1),t} \neq \mathrm{b}^{f(G_2),t}$. Furthermore, from Equation \eqref{color}, we have that if $\mathrm{a}^{f(G_1)^{(1)},t} \neq \mathrm{a}^{f(G_2)^{(1)},t}$, then $\mathrm{c}^{G_1,t} \neq \mathrm{c}^{G_2,t}$. By transitivity, it follows that if $\mathrm{c}^{G_1,t} \neq \mathrm{c}^{G_2,t}$, then $\mathrm{b}^{f(G_1),t} \neq \mathrm{b}^{f(G_2),t}$, implying that for any pair of non-isomorphic graphs distinguishable by the $\text{WL}$ coloring $\mathrm{c}$, the $\text{CCNN}$ coloring $\mathrm{b}$ can also distinguish them.

The next goal is to prove $\mathrm{a}^t\preceq \mathrm{b}^t$. We now proceed by induction on the iteration $t$.

\textbf{1) Base Case.} At iteration $t=0$, the initial features (or colours) of the 0‐cells in $\text{CCNN}$ are obtained directly from the node features of the graph. By construction, we have $\mathrm{b}^{f(G_1),0}= \mathrm{b}^{f(G_2),0} \Longrightarrow \mathrm{a}^{G_1,0}= \mathrm{a}^{G_2,0}$.

\textbf{2) Inductive Step.} Assume that after $t$ iterations the $\text{WL}$ colouring $\mathrm{a}^t$ is topological reduced to the $\text{CCNN}$ colouring $\mathrm{b}^t$, i.e., $\mathrm{a}^t\preceq \mathrm{b}^t$. Consider two 0‐cells $\sigma$ in $X$ and $\tau$ in $Y$ such that $\mathrm{b}^{t+1}(\sigma)=\mathrm{b}^{t+1}(\tau)$. We know that
\begin{equation}
    \mathrm{b}^{t+1}(\sigma)=\{\!\!\{\mathrm{b}^{t}(\sigma), \mathrm{b}^{t}(\mathcal{B}(\sigma)), \mathrm{b}^{t}(\mathcal{C}(\sigma)), \mathrm{b}^{t}(\mathcal{N}_{\downarrow}(\sigma)), \mathrm{b}^{t}(\mathcal{N}_{\uparrow}(\sigma))\}\!\!\},
\end{equation}
However, for 0-cells $\sigma$, the boundary $\mathcal{B}(\sigma)$ and lower adjacency $\mathcal{N}_{\downarrow}(\sigma)$ are not defined, so we only need to consider $\mathrm{b}^{t}(\sigma)$, $\mathrm{b}^{t}(\mathcal{C}(\sigma))$, and $\mathrm{b}^{t}(\mathcal{N}_{\uparrow}(\sigma))$. Given that $\mathrm{b}^{t+1}(\sigma)=\mathrm{b}^{t+1}(\tau)$, we obtain:
\begin{equation}
    \mathrm{b}^{t}(\sigma)=\mathrm{b}^{t}(\tau), \ \mathrm{b}^{t}(\mathcal{C}(\sigma))=\mathrm{b}^{t}(\mathcal{C}(\tau)) \ \text{and} \ \mathrm{b}^{t}(\mathcal{N}_{\uparrow}(\sigma))=\mathrm{b}^{t}(\mathcal{N}_{\uparrow}(\tau)).
\end{equation}
By the inductive hypothesis $\mathrm{a}^t\preceq \mathrm{b}^t$, we further obtain $\mathrm{a}^{t}(\sigma)=\mathrm{a}^{t}(\tau)$ and
\begin{equation}
\begin{aligned}
       &\mathrm{a}^{t}(\mathcal{N}_{\uparrow}(\sigma))=\{\!\!\{\mathrm{a}^{t}(\delta_1)| \delta_1 \in \mathcal{N}_{\uparrow}(\sigma) \}\!\!\} \\
     =&\{\!\!\{\mathrm{a}^{t}(\delta_2)| \delta_2 \in \mathcal{N}_{\uparrow}(\tau) \}\!\!\}=\mathrm{a}^{t}(\mathcal{N}_{\uparrow}(\tau)). \label{eq1}  
\end{aligned}
\end{equation}

Since the 1-skeleton corresponds to a graph, and $\text{WL}$ coloring aggregates only self and neighbor features, we have $\mathrm{a}^{t+1}(\sigma)=\{\!\!\{\mathrm{a}^{t}(\sigma),  \mathrm{a}^{t}(\mathcal{N}_{\uparrow}(\sigma))\}\!\!\}$. Combining this with Equation \eqref{eq1}, we conclude that $\mathrm{a}^{t+1}(\sigma)=\mathrm{a}^{t+1}(\tau)$. Finally, since $\mathrm{b}^{t+1}(\sigma)=\mathrm{b}^{t+1}(\tau)$ implies $\mathrm{a}^{t+1}(\sigma)=\mathrm{a}^{t+1}(\tau)$, we establish that $\mathrm{a}^{t+1}\preceq \mathrm{b}^{t+1}$.

By the inductive hypothesis, we have proven that $\mathrm{a}^t\preceq \mathrm{b}^t$. Further leveraging transitivity and Theorem \ref{theorem1}, we conclude that $\mathrm{c}^t\preceq \mathrm{b}^t$. \hfill $\square$

\subsection{Causal Background Knowledge}
\label{causal}

\textbf{Structural Causal Models and Intervention.} A structural causal model (SCM)~\cite{pearl2009causality} is a triple $M=\left \langle X,U,F \right \rangle$, where $U$ is known as the \textit{exogenous variable}, determined by external factors of the model. $X=\left \{ X_1,X_2,...,X_n \right \}$ is referred to as the \textit{endogenous variable}, whose changes are determined by the functions $F=\left \{ f_1,f_2,...,f_n \right \}$. Each $f_i$ represents $\left \{ f_i:U_i\cup PA_i \rightarrow X_i\right \}$, where $U_i\subseteq U$, $PA_i\subseteq X\backslash X_i$, satisfying:
\begin{equation}
    x_i=f_i\left ( pa_i,u_i \right ),\quad i=1,2,...,n.
\end{equation}
Each causal model $M$ corresponds to a directed acyclic graph (DAG) $G$, where each node corresponds to a variable in $X\cup U$, and directed edges point from $U_i\cup PA_i$ to $X_i$.

An \textit{intervention} refers to forcing a variable $X_i$ to take a fixed value $x_i$. This equivalently removes $X_i$ from the influence of its original functional mechanism $x_i=f_i\left ( pa_i,u_i \right )$ and replaces it with a constant function $X_i=x_i$. Formally, we denote the intervention as $do(X_i=x_i)$, or simply $do(x_i)$. After the intervention on $X_i$, the corresponding causal graph $G_{x_i}$ is obtained by removing all arrows pointing to $X_i$ in $G$ to represent the post-intervention world.

\textbf{Path and $d$-separation.} We summarize definitions~\cite{pearl2009causality} to help us determine the independence between variables in the SCM graph.

\begin{definition}
    \label{path} \textbf{(Path)}
    In the SCM graph, the paths from variable $X$ to $Y$ include three types of structures: 1) Chain Structure: $A \rightarrow B\rightarrow C$ or $A \leftarrow B\leftarrow C$, 2) Fork Structure: $A \leftarrow B\rightarrow C$, and 3) Collider Structure: $A \rightarrow B\leftarrow C$.
\end{definition}

\begin{definition}
    \label{separation} \textbf{($d$-separation)}
A path $p$ is blocked by a set of nodes $Z$ if and only if:
\begin{enumerate}
    \item $p$ contains a chain of nodes $A \rightarrow B\rightarrow C$ or a fork  $A \leftarrow B\rightarrow C$ such that middle node $B$ is in $Z$ (i.e., $B$ is conditioned on), or
    \item $p$ contains a collider $A \rightarrow B\leftarrow C$ such that the collider node $B$ is not in $Z$, and no descendant of $B$ is in $Z$. 
\end{enumerate}
\end{definition}
If $Z$ blocks every path between two nodes $X$ and $Y$ , then $X$ and $Y$ are \textit{$d$-separated}, conditional on $Z$, and thus are independent conditional on $Z$, denoted as $X \upmodels Y \mid Z$.

\textbf{Backdoor and Backdoor Adjustment.} 
\begin{definition}
    \label{Back-Door} \textbf{(Backdoor)}
In a DAG $G$, a set of variables $Z$ satisfies the backdoor criterion for an ordered pair of variables $(X_i, X_j)$ if:
\begin{enumerate}
    \item No node in $Z$ is a descendant of $X_i$.
    \item $Z$ blocks all paths between $X_i$ and $X_j$ that are directed into $X_i$.
\end{enumerate}
Similarly, if $X$ and $Y$ are two disjoint subsets of nodes in $G$, then $Z$ is said to satisfy the backdoor criterion for $(X, Y)$ if $Z$ satisfies the backdoor criterion for any pair of variables $(X_i, X_j)$, where $X_i \in X$ and $X_j \in Y$.
\end{definition}

\begin{definition}
    \label{adjustment} \textbf{(Backdoor adjustment)}
If a set of variables $Z$ satisfies the backdoor criterion for $(X, Y)$, then the causal effect of $X$ on $Y$ is identifiable and can be given by the following formula:
\begin{equation}
    P(Y=y\mid do(X=x))=\sum _{z}P(Y=y\mid X=x,Z=z)P(Z=z).
\end{equation}
\end{definition}

\begin{table}[t]
\setlength{\tabcolsep}{1.8pt}
\begin{center}
\begin{tabular}{ccc}
\toprule
\multirow{2}{*}{Experiment} & \multicolumn{2}{c}{Learning Type}  \\  
\cmidrule(lr){2-3} & Unsupervised & Semi-supervised \\  
\midrule
Number of layers & 3 & 3 \\
Cellular embedding dim & 32 & 32 \\
Number of projections & 2 & 2 \\
Nonlinear transformation dim & 96 & 96\\
Graph norm & BatchNorm & BatchNorm \\
Jump Mode & cat & cat \\
Pooling & global\_add\_pool & global\_add\_pool \\
Pre-train $lr$ & 0.001 & 0.001\\
Temperature $\tau$ & 0.2 & 0.2\\
Training epochs & 20 & 20 \\
Batch size & 128 & 128 \\
Permuted rate $\eta$ & \{0.1,1,10,100\} & \{0.1,1,10,100\} \\
Ring size $k$ & 6 & 6 \\
\bottomrule
\end{tabular}
\end{center}
\caption{Model architectures and hyper-parameters.}
\vspace{-0.6cm}
\label{tab:model architectures and hyper-parameters}
\end{table}

\section{More Experimental Details}

\subsection{Model Configurations}
The details of our model architectures and corresponding hyper-parameters are summarized in Table \ref{tab:model architectures and hyper-parameters}.

\subsection{Running Environment}
The results of unsupervised learning and semi-supervised learning were obtained using a single NVIDIA V100 GPU with 32G of memory. We performed the experiments on Ubuntu 20.04 as our operating system.

\subsection{Additional Hyper-parameter Analysis}

\begin{figure}[ht]
\vspace{-0.2cm}
    \centering
    \includegraphics[width=1.0\linewidth]{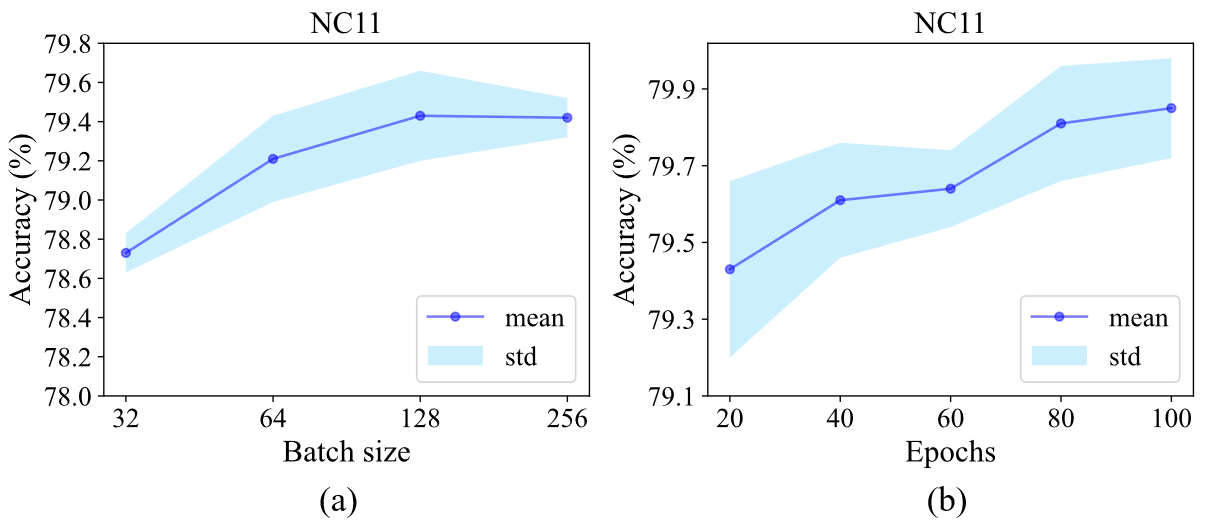}
    \vspace{-0.6cm}
    \caption{Hyper-parameter sensitivity analysis.}
    \label{fig:hyper-batch-epoch}
    \vspace{-0.45cm}
\end{figure}

\textbf{Batch size.} Figure \ref{fig:hyper-batch-epoch}(a) shows the classification accuracy of our models after training twenty epochs using different batch sizes from 32 to 256 on the NCI1 dataset. We observe that increasing the batch size led to improved model performance. 

\textbf{Epochs.} Figure \ref{fig:hyper-batch-epoch}(b) illustrates the classification accuracy on the NCI1 dataset across different training epochs, ranging from 20 to 100. We observe a general upward trend in accuracy as the number of epochs increases, indicating that prolonged training allows the model to learn more discriminative representations. Additionally, the variance remains relatively stable, suggesting the robustness of our approach across different training durations.

\begin{figure}[ht]
    \centering
    \includegraphics[width=0.65\linewidth]{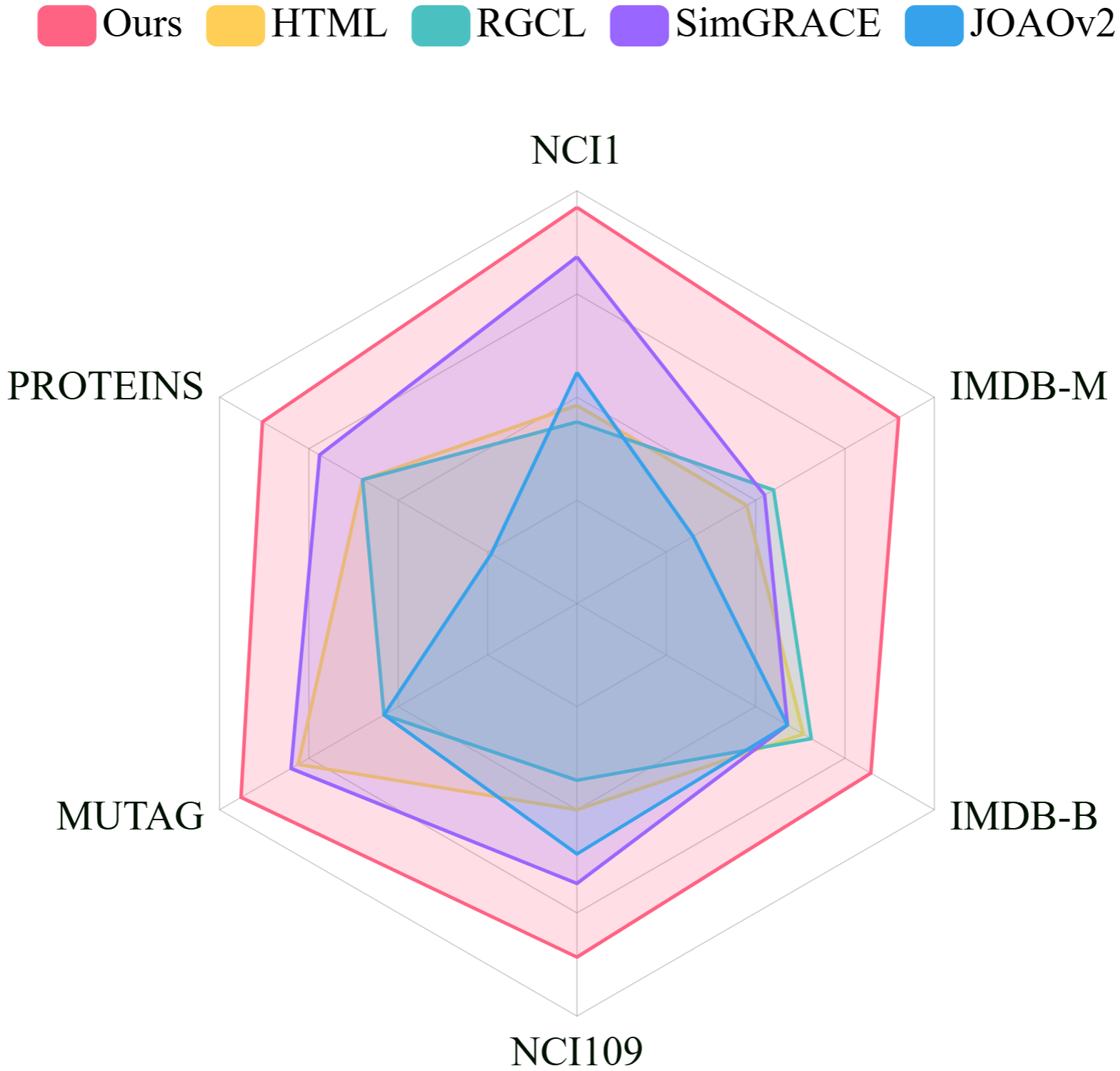}
    \vspace{-0.3cm}
    \caption{Visualization of unsupervised learning results.}
    \label{fig:radar}
\end{figure}

\subsection{Additional Visualization Results}
Figure \ref{fig:radar} illustrates the results of unsupervised learning comparisons through a radar chart, where each axis represents a dataset, and the vertices correspond to the classification accuracy of different methods. The different colors denote the top-5 performing models. Our method consistently outperforms others across multiple datasets, highlighting its robustness and generalization ability in unsupervised representation learning.

\end{document}